\documentclass[a4paper]{cas-dc}
\usepackage[numbers,sort&compress]{natbib}
\usepackage{algorithmic}
\usepackage{algorithm}

\def\tsc#1{\csdef{#1}{\textsc{\lowercase{#1}}\xspace}}
\tsc{WGM}
\tsc{QE}

\begin{document}
\let\WriteBookmarks\relax
\def\floatpagepagefraction{1}
\def\textpagefraction{.001}
\let\printorcid\relax 

\shorttitle{}    

\shortauthors{Yuhan Kang et al.}

\title [mode = title]{SurANet: Surrounding-Aware Network for Concealed Object Detection via Highly-Efficient Interactive Contrastive Learning Strategy}                      

\author[1]{Yuhan Kang}
\credit{Writing – review \& editing, Writing – original draft, Methodology, Software}
\ead{kyh433@hnu.edu.cn}

\affiliation[1]{organization={The College of Electrical and Information Engineering, Hunan University},
                city={Changsha},
                country={China}}

\author[2]{Qingpeng Li}
\cormark[1]
\ead{liqingpeng@hnu.edu.cn}
\credit{Writing – review \& editing, Funding acquisition, Conceptualization, Data curation}

\affiliation[2]{organization={The School of Robotics, Hunan University},
                city={Changsha},
                country={China}}

\author[1]{Leyuan Fang}
\credit{Writing – review \& editing, Supervision}
\ead{fangleyuan@gmail.com}

\author[3,4]{Jian Zhao}
\credit{Writing – review \& editing, Supervision}
\ead{jian_zhao@nwpu.edu.cn}

\affiliation[3]{organization={China Telecom Institute of AI},
    city={Beijing},
    country={China}}

\affiliation[4]{organization={Northwestern Polytechnical University},
                city={Xi'an},
                country={China}}

\author[1,3]{Xuelong Li}
\credit{ Supervision}
\ead{li@nwpu.edu.cn}
\affiliation[5]{organization={China Telecom},
    city={Beijing},
    country={China}}

\affiliation[6]{organization={the School of Artificial Intelligence, Optics and Electronics (iOPEN), Northwestern Polytechnical University},
                city={Xi'an},
                country={China}}

\cortext[1]{Corresponding author: Qingpeng Li.}  
\begin{abstract}
    Concealed object detection (COD) in cluttered scenes is significant for various image processing applications. However, due to that concealed objects are always similar to their background, it is extremely hard to distinguish them. Here, the major obstacle is the tiny feature differences between the inside and outside object boundary region, which makes it trouble for existing COD methods to achieve accurate results. In this paper, considering that the surrounding environment information can be well utilized to identify the concealed objects, and thus, we propose a novel deep Surrounding-Aware Network, namely SurANet, for COD tasks, which introduces surrounding information into feature extraction and loss function to improve the discrimination. First, we enhance the semantics of feature maps using differential fusion of surrounding features to highlight concealed objects. Next, a Surrounding-Aware Contrastive Loss is applied to identify the concealed object via learning surrounding feature maps contrastively. Then, SurANet can be trained end-to-end with high efficiency via our proposed Spatial-Compressed Correlation Transmission strategy after our investigation of feature dynamics, and extensive experiments improve that such features can be well reserved respectively. Finally, experimental results demonstrate that the proposed SurANet outperforms state-of-the-art COD methods on multiple real datasets. Our source code will be available at \textrm{\url{https://github.com/kyh433/SurANet}}
\end{abstract}



\begin{keywords}
    Concealed Object Detection \sep Surrounding-Aware Analysis \sep Efficient Contrastive Learning \sep Deep Neural Network
\end{keywords}

\maketitle

\section{Introduction}
\label{sec:introduction}
Concealed object detection (COD) is an attractive and significant task to identify objects which often conceal their texture in the surrounding environment~\cite{le2019anabranch,fan2023advances}. 
Since the potential value of concealed objects, COD has been widely used in various applications, such as species protection~\cite{fan2021concealed, perez2012early}, emergency rescue~\cite{lygouras2019unsupervised}, national defense~\cite{liu2021camouflaged, hu2020infrared}, medical diagnosis~\cite{ji2022fast, fan2020pranet, mangal2020covidaid}, and industry production. 
In natural scenes, background matching and disruptive camouflage are two common concealing methods, in which concealed objects aim to minimize their own signals via adjust their color, shape or texture to the environment~\cite{hughes2019imperfect}. As a result, concealed objects can be easily hidden in the environment, which makes the detection and identification task always difficult.

\begin{figure}[tb!]
	\centering
	\includegraphics[width=85mm]{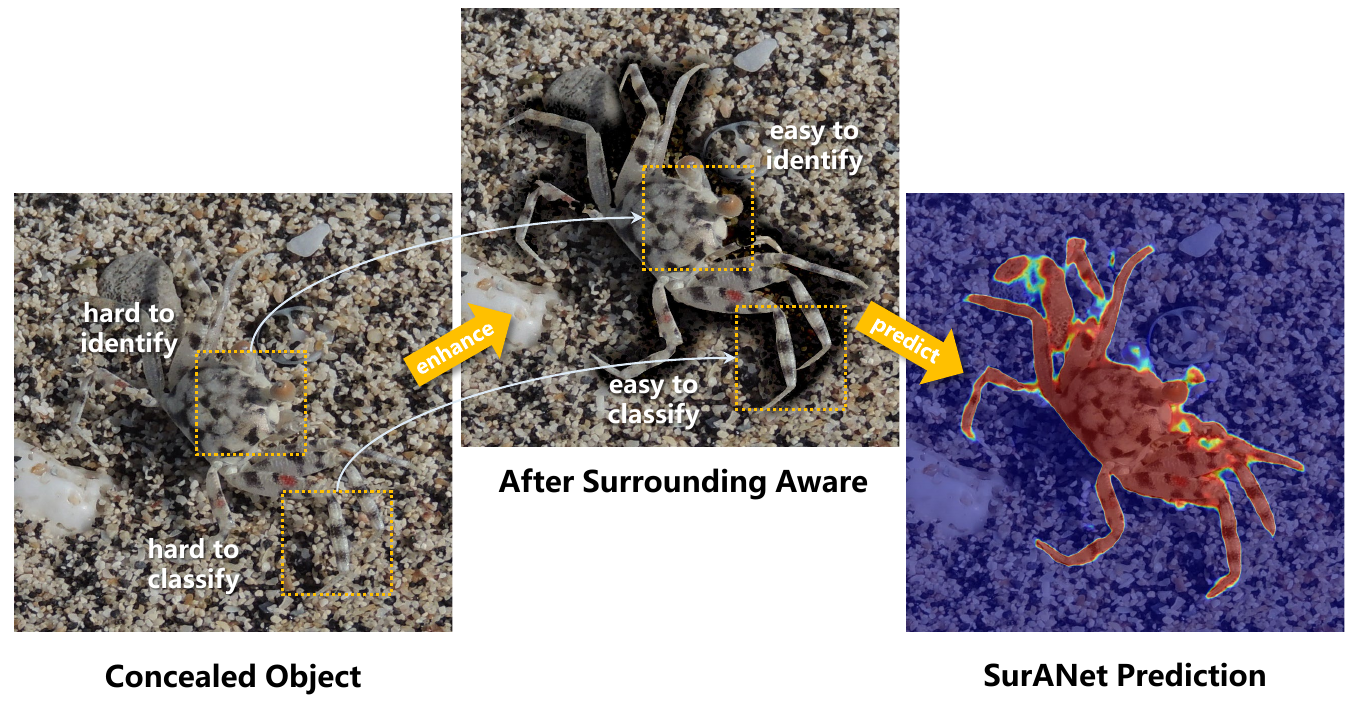}
	\caption{\textrm{Surrounding information can be aided for detecting concealed objects. The left picture contains a concealed sand bubble crab which is difficult to identify. After noticing its around surrounding area, it is easier to identify the object and classify the details.}}
	\label{fig1}
\end{figure}

With the significant progress in the COD research development, many feature extractors~\cite{bi2021rethinking, yang2021uncertainty,jia2022segment,huang2023feature,ZHOU2024128395_Neuro,song2023fsnet, LIANG2024127050_Neuro} are designed to improve the state-of-the-art (SOTA) performance. Especially in recent years, inspired by the hunting process of predators, Fan et al.~\cite{fan2020camouflaged} construct a search-identified feature extractor by enhancing the texture feature of concealed objects several times. 
Thereafter, Ren et al.~\cite{ren2021deep} propose multiple texture-aware refinement modules to enhance concealed object texture features. Zhong et al.~\cite{zhong2022detecting} design a frequency feature extractor to better obtain object texture features in frequency domain. Zhu et al.~\cite{zhu2021inferring} design an interactive feature extractor based on the complementary relationship between the boundary feature and the texture feature. Overall, such feature extractors concentrate particular emphasis on the texture, edge, and frequency feature of concealed objects, and are designed with the focus on the characteristics of concealed objects.

Meanwhile, the design of classifier and loss function is becoming increasingly important for COD~\cite{bi2021rethinking, LI2023126530_Neuro,hu2023high,song2023pixel}. For example, Pang et al.~\cite{pang2022zoom} design a strong constraint uncertainty-aware loss as the auxiliary of binary cross-entropy loss, which enhances the decision confidence of texture features and increases the penalty for a fuzzy prediction. Li et al.~\cite{li2021uncertainty} introduce a novel similarity measurement to enhance the detection ability of concealed objects. The aforementioned studies focus on learning binary differences between concealed objects and background, revealing that more effective feature representations can acquire better detection results. However, when the texture of concealed objects closely resembles their surrounding area, detection tasks become more challenging. Therefore, the main challenge of COD tasks is that, how to classify the tiny differences between the inside and outside of the concealed object boundary regions with acceptable computing efficiency.

In nature, the maintenance of concealing are expensive~\cite{ruxton2018avoiding}. When concealed objects cannot match their surrounding environment, they are more likely to be discovered. Thus, as shown in Fig.~\ref{fig1}, the main motivation of this work, is to use surrounding area features for chopping off the texture consistency between the concealed object and its background. Here, when we focus on the surrounding area, the object detail will be noticed in a smaller search domain after the differential representation of surrounding features, which will make it easier to spot and identify concealed objects.

Motivated by the observation above, in this paper, we propose a novel deep Surrounding-Aware Network (namely SurANet) which holistically feeds the surrounding area information into the object feature extractor and classifier. First, we design a novel Surrounding-Aware Enhancement module, not only to enhance the surrounding textures, but also to improve the object textures layer by layer correspondingly. Next, the surrounding difference contrast loss is utilized to further identify the feature difference between the concealed object and the background by comparing the surrounding area. The experimental results reveal that our loss performs well without adding excessive computational burden. Benefiting from the design of this method, SurANet significantly improves the identification ability of concealed objects, and achieving SOTA on all existing COD datasets.

\textcolor{black}{In summary, this paper makes the following three main contributions:}

\begin{itemize}
	\item[$\bullet$] \textcolor{black}{Firstly, we discovered that focusing more on the surrounding area could contribute to distinguishing concealed objects, and experimental validation demonstrates the impact of surrounding awareness of segmenting, which provides a novel solution for COD task.}
	
	\item[$\bullet$] \textcolor{black}{Secondly, we proposed a novel deep neural network called Surrounding-Aware Network (SurANet), including Surrounding-Awareness Enhancement Module and Surrounding-Awareness Contrastive Loss function, enhancing the features between concealed object and surrounding area, respectively, and achieving effective awareness of concealed object.}
	
	\item[$\bullet$] \textcolor{black}{Thirdly, we proposed a novel training strategy, called Spatial-Compressed Correlation Transmission (SCCT), based on the investigation of feature dynamics. Via extensive experiments, it indicates that SCCT can achieve optimal performance for COD task without introducing significant additional computational costs, which provides valuable capacity for practical application.}
\end{itemize}

The remaining parts of this paper are organized as follows. In Section II, we review several mainstream methods for COD research. In Section III, we introduce the SurANet in detail. In Section IV, we demonstrate the efficiency of SurANet through performance evaluation and ablation experiments. Finally, the conclusion is given in Section V.

\section{Related Work}
\label{sec:relatedwork}
In previous works~\cite{fan2020camouflaged, bi2021rethinking}, COD methods can be generally categorized into tow main steps: the feature extractor and classifier. And hence, researchers focus on two key issues: 1) how to effectively extract concealed semantic relationships from sparse vectors, 2) how to efficiently use semantic information to stretch the distance between concealed objects and background for better discrimination. Next, we will review the related works of such above issues. 

\begin{figure*}[tb!]
	\setlength{\abovecaptionskip}{0cm} 
	\setlength{\belowcaptionskip}{-0.2cm}
	\centering
	\includegraphics[width=170mm]{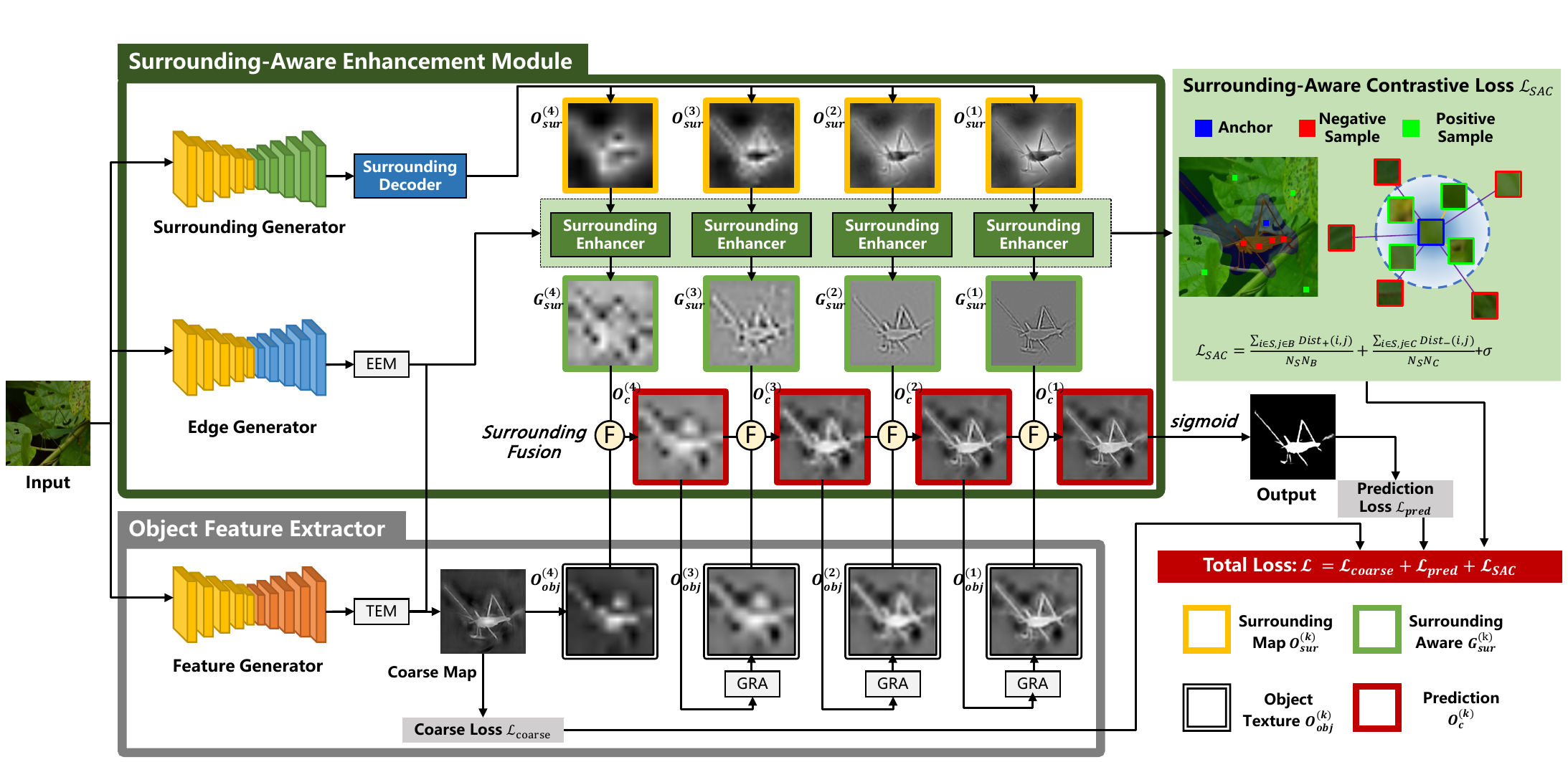}
	\caption{\textrm{Overall architecture of Surrounding-Aware Network (SurANet). We proposed Surrounding-Aware Enhancement module and designed Surrounding-Aware Contrastive Loss to further enhance the surrounding awareness. Specifically, the input image passes through three different generators to obtain object features of each layer, including texture features, surrounding features, and edge features. Subsequently, the network enhances surrounding features layer by layer, through surrounding enhancer and surrounding fusion. Finally, under the supervision of Surrounding-Aware Contrastive Loss, the network progressively refines segmentation results of concealed objects through surrounding fusion. The details of edge enhance module (EEM) and texture enhance module (TEM) are implemented in~\cite{fan2021concealed}.}}
	\label{fig2}
\end{figure*}

\subsection{Feature Extractor in COD}
\label{subsec:Extractor}
\textcolor{black}{The early work of COD feature extractor mainly focus on handcrafted operators~\cite{stevens2009animal}.} For instance, Sengottuvelan et al.~\cite{sengottuvelan2008performance} recognize the concealed objects via a co-occurrence matrix. Pan et al.~\cite{pan2011study} propose a 3D convexity model to detect concealed objects. \textcolor{black}{Due to concealed objects always contain more confusing textures, such above common feature extractors could not perform so well. In recent years, COD feature extractors mainly adopt enhancement and fusion strategies according to the object feature.} For example, Fan et al.~\cite{fan2020camouflaged} propose SINet based on the enhancement strategy. Here, they design different receptive field components to extract coarse maps containing object semantic information, and in advance, to further enhance the object feature via such coarse maps. 

In subsequent, Fan et al.~\cite{fan2021concealed} optimize the feature extractor and propose SINetV2. 
They design texture feature enhancement module which is used to capture fine-grained texture features with local-global contextual, and the proposed group-reversal attention block can improve the detail of concealed objects. 

Moreover, Ren et al.~\cite{ren2021deep} propose multiple texture-aware refinement modules to improve concealed object texture features, via amplifying subtle texture differences between the object and background. 
Zhong et al.~\cite{zhong2022detecting} introduce the frequency domain information, and design a frequency enhancement module, which can increase the discrimination of concealed objects. 
Jiang et al.~\cite{10224812_TIM} propose a JCNet, which introduce salient object features to effectively enhance the concealed object extraction capability. 

On the other side, as multiple concealed object features can be considered, feature fusion is improved to be a better way. For example, TINet is proposed based on an interactive guidance framework, focusing on searching the concealed object boundary and texture differences via interactive feature fusion strategy~\cite{zhu2021inferring}. 
Lv et al.~\cite{lv2021simultaneously} propose RankNet to simultaneously localize, segment and rank concealed objects, fusing the general texture features and specific features through an channel attention module and an position attention module, respectively. What's more, Zhai et al.~\cite{zhai2021mutual} propose a mutual graph learning model. Here, the feature extractor locates the concealed object features through circular reasoning strategy, and such features are subsequently fused together with boundary features for mutual benefit.

\textcolor{black}{Overall, on the one hand, the feature enhancement methods make it easier for feature extractor to identify potential object areas. On the other hand, the feature fusion methods enable feature extractors to achieve consistent representations across different dimensions of concealed objects. However, current feature extractors are based on the texture, edge, and frequency domain characteristics of concealed objects, ignoring their surrounding environment, and such oversight makes it challenging to detect details when the object resembles their surroundings closely.}

\subsection{Classifier and Loss Function in COD}
\label{subsec:Classifier}
The design of the classifier and loss function is particularly critical for COD tasks~\cite{he2022weakly, liu2021integrating, liu2022modeling}. The most common approach is to use cross-entropy loss $\mathcal{L}_{CE}$, defined as:
\begin{small}
	\begin{equation}
		\begin{split}
			\mathcal{L}_{CE}&=-\frac{1}{N}\sum_{(i, j)}[GT(i, j)log(O(i, j)) \\
			& +(1-GT(i, j))log(1-O(i, j))].
		\end{split}
	\end{equation}
\end{small}

\noindent where $GT(i, j)\in{0, 1}$ and $O(i, j)\in[0, 1]$ denote ground truth and predicted value in position $(i, j)$, respectively. While considering multiple features of concealed objects, different $\mathcal{L}_{CE}$ are always computed by weighted stack~\cite{liu2021integrating, park2022tcu}. In recent years, according to the characteristics of concealed objects, a series of works have designed the corresponding classifiers. For example, Pang et al.~\cite{pang2022zoom} design a strong constraint uncertainty-aware loss, formulated as $\mathcal{L}_{UAL}=1- \left|2O(i, j)-1\right|^{2}$, and based on this, they modify $L_{CE}$ to enhance the decision confidence of texture features, and increase the penalty of fuzzy prediction. What's more, Li et al.~\cite{li2021uncertainty} feed simple positive samples from a COD dataset to a salient object detection (SOD) model, and introduce a similarity loss $\mathcal{L}_{latent}$ to measure the difference between saliency feature $f_{sp}$ and concealed feature $f_{cp}$ in latent space. Here, $\mathcal{L}_{latent}$ is defined as:
\begin{small}
	\begin{equation}
		\mathcal{L}_{latent}=cos(f^{sp}, f^{cp})=\frac{f^{sp}\cdot f^{cp}}{||f^{sp}||\times ||f^{cp}||}.
	\end{equation}
\end{small}

\noindent Liu et al.~\cite{liu2021confidence} present a confidence-aware camouflaged object detection framework using dynamic supervision, which introduces a confidence-aware learning strategy to pay more attention to the hard samples in the loss function. Above all, these studies classify concealed objects by learning binary differences between objects and their background, and to achieve fine segmentation result always requires large computational resources. Despite the growing interest, COD research has not yielded satisfactory results in classification details.

Contrastively, our proposed work differs from existing research in the following two aspects.

1) In terms of feature extraction, different from the previous COD work focusing more on the object feature, we propose a surrounding-aware perspective for feature extractor to better distinguish the concealed object details. With the enhancement and fusion method applied to the surrounding feature, it makes more easier to classify the details of concealed objects.

2) In terms of classifier, we introduce a comparative loss function to efficiently differ between the surrounding area and concealed objects. Existing works always learn binary differences between concealed objects and their background, we utilize a triple comparison involving concealed objects, surrounding area, and background. Instead of a complex pixel-wise contrastive loss, we significantly improve learning efficiency without costly computational expenses by SCCT strategy.

\begin{figure*}[tb!]
  \setlength{\abovecaptionskip}{0cm} 
  \setlength{\belowcaptionskip}{-0.2cm}
  \centering
  \includegraphics[width=170mm]{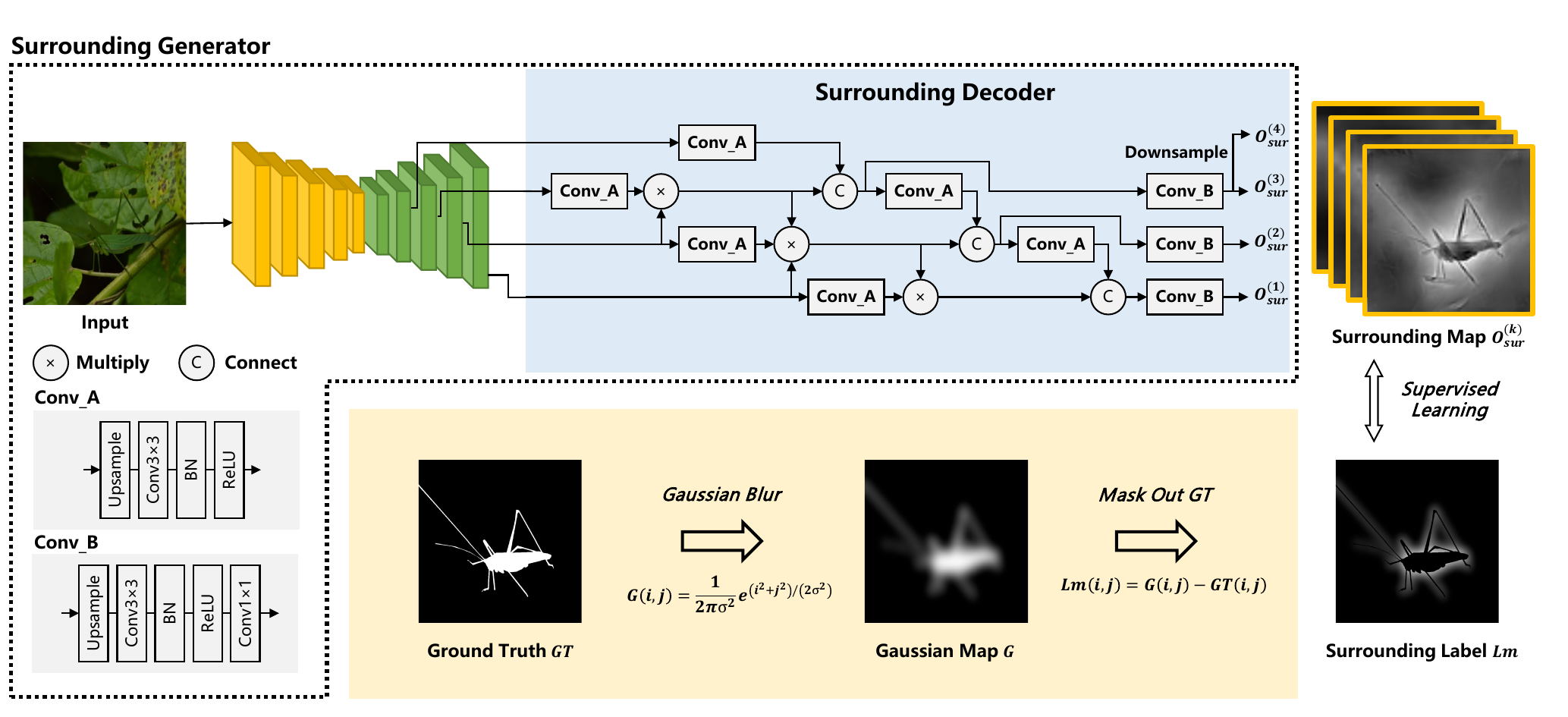}
  \caption{\textrm{\textcolor{black}{Surrounding Generator of SAE module. The surrounding label is composed of the periphery of ground truth after Gaussian blurring. Under hierarchical computation of surrounding decoder, the network generates surrounding maps on various spatial scales, effectively representing the surrounding environment.}}}
  \label{fig3}
\end{figure*}

\section{Methodology}
\label{sec:methodology}
In nature, as shown in Fig.~\ref{fig1}, concealed objects are more likely to be discovered when their camouflage can not match the surrounding environment. 
Inspired by such point, we focus on capturing the network awareness of the surrounding environment area for guiding the learning process better. 
In this section, we provide a detailed introduction to SurANet.

\subsection{Network Architecture}
\label{subsec:NetArch}
The overall implementation framework of the SurANet is shown in Fig.~\ref{fig2}. SurANet mainly consists of an cascaded extractor and a classifier. Here, the extractor is responsible for key features of the concealed objects, and the classifier efficiently discriminates the extracted features, respectively. On the one hand, for a better distinction between concealed objects and there surrounding environment features, we introduce a Surrounding-Aware Enhancement module (namely SAE) for the object feature extractor, to further enhance the awareness of the surrounding environment around concealed objects. On the other hand, in the design of the classifier, we use Surrounding-Aware Contrastive Loss function (namely SACLoss), which makes the difference between surrounding feature and concealed object feature more pronounced. Next, we will introduce such aforementioned parts in detail.

\begin{figure}[tb!]
	\setlength{\abovecaptionskip}{0cm} 
	\setlength{\belowcaptionskip}{-0.2cm}
	\centering
	\includegraphics[width=85mm]{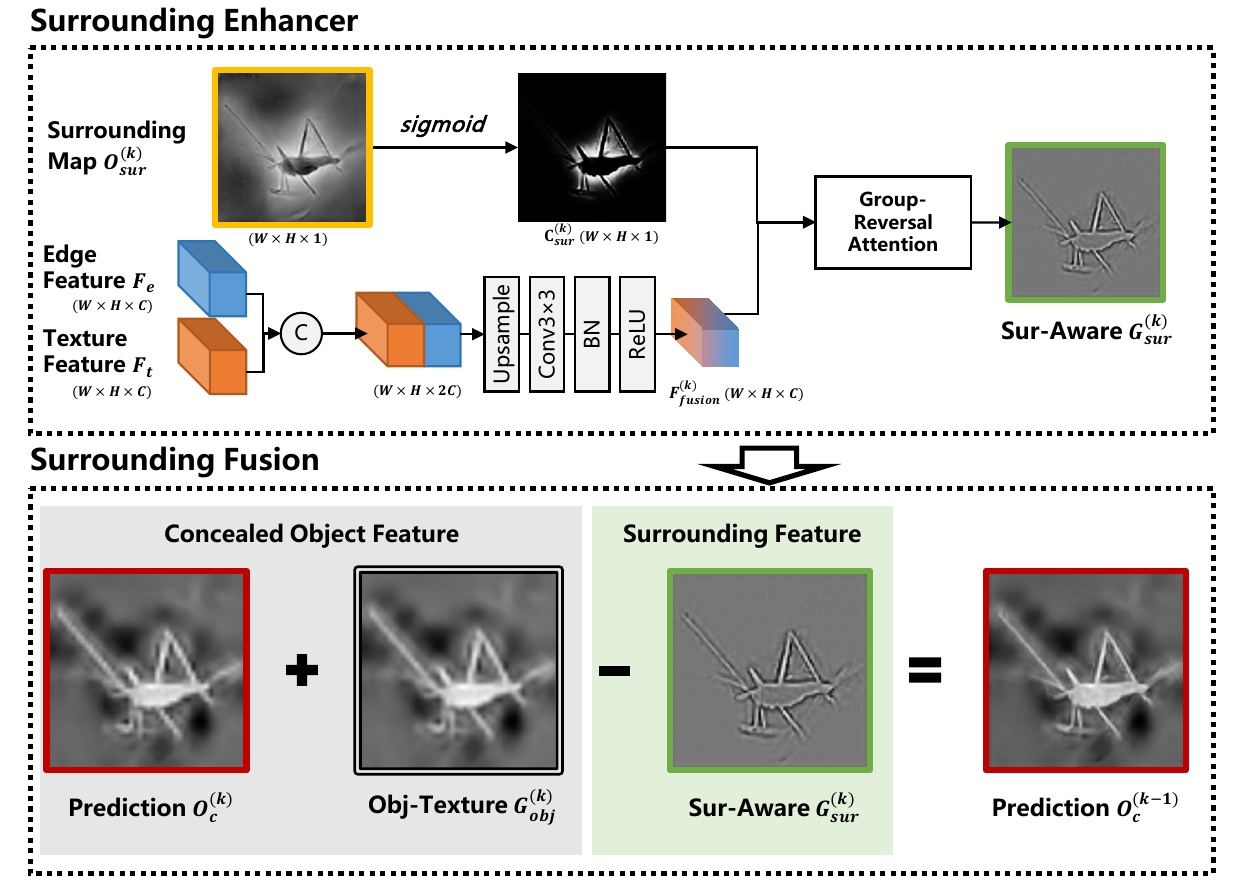}
	\caption{\textrm{\textcolor{black}{Surrounding enhancer and surrounding fusion in SAE module. Surrounding-aware feature (Sur-Aware $G_{sur}$) is composed of texture feature $F_t$ and edge feature $F_e$ under the constraint of surrounding map. By iteratively fusing texture feature (Obj-Texture $G_{obj}$ ) and surrounding-aware feature on coarse prediction (predicted $O_c$), SurANet enhances the awareness of object features and surrounding features. Therefore, the final prediction effectively reduces environmental noise and enhances object details. The detail of texture feature and group-reversal attention (GRA) are implemented in~\cite{fan2021concealed}.}}}
  \label{fig4}
  \end{figure}

\subsection{Surrounding-Aware Enhancement Module}
\label{subsec:SAEModule}
In SurANet, object feature extractor (OFE) is mainly responsible for extracting features of concealed objects. The initial segmentation result (so-called coarse map) of concealed objects, can be obtained through OFE as shown in Fig.~\ref{fig2}. However, when the camouflage closely resembles the surrounding environment, only relying on OFE for feature extraction is not sufficient. Thus, we propose SAE for better perception of the surrounding environment. SAE is mainly divided into two parts, surrounding generator and surrounding enhancer. Here, as shown in Fig.~\ref{fig3}, in order to obtain effective surrounding area information, we first design the feature representation of local environment information around the concealed objects and build a surrounding generator. 

On the one hand, since the semantic relationship between various concealed objects and corresponding background is essentially different (e.g. animals disguised as leaves, twigs, gravel, rocks, etc.), the surrounding area always does not contain clear range. On the other hand, the similarity between environment and concealed objects texture, gradually decreases as the surrounding area expands. In this case, the surrounding labels created by Gaussian blur~\cite{gedraite2011investigation} can effectively convey the knowledge of different background areas, and meanwhile, reduce the knowledge of the background away from the object. Hence, it can well describe the surrounding environmental information in this way, which aids in distinguishing the subtle differences between objects and background with more segmentation precise.

For simplicity, we apply Gaussian Blur for binary segmentation labels of concealed objects to obtain surrounding information, and mask out the location of concealed objects to preserve the surrounding area only, respectively. Via hierarchical computation in surrounding Decoder, the proficiency of network for discerning surrounding features across various scales, is significantly augmented. Such process culminates in the generation of different-scaled surrounding maps, effectively representing the ambient environment naturally.

Formally, let $I\in\mathbb{R}^{C\times H\times W}$ be the input image, where $C, H, W$ denote the number of channels, height, and width, respectively. $C=3$. Similarly, let $GT\in \mathbb{R}^{H\times W}$ be the ground truth of the input image. We first apply Gaussian blur on $GT$, namely:
\begin{small}
\begin{equation}
  Gaus_{\sigma}=\frac{1}{2\pi\sigma^2}e^{(i^2+j^2 )/(2\sigma^2)}
\end{equation}
\end{small}
\begin{small}
\begin{equation}
  G(i,j)=GT \ast Gaus_{\sigma},
\end{equation}
\end{small}

\noindent where $G(i, j)$ denotes the value of corresponding pixels after Gaussian Blur $Gaus_{\sigma}$. $\sigma$ denotes the variance which determines the size of surrounding area. Then, we just zero out the location of concealed objects to preserve surrounding area $Lm$ only as follows:
\begin{small}
\begin{equation}
  Lm(i, j)=max\left \{G(i, j)-GT(i, j), 0\right \}. 
\end{equation}
\end{small}

After the representation of surrounding area, we can extract surrounding feature $F_{sur}$ from Encoder $E$, and generate surrounding map $O_{sur}$ from Decoder $D$, namely:
\begin{small}
\begin{equation}
  O_{sur}=D(F_{sur})=D(E(I)).
\end{equation}
\end{small}

However, just using the surrounding generator to produce surrounding maps is insufficient. \textcolor{black}{Since erroneous pixels always lead to accumulative errors, which eventually brings the missing of concealed objects, we will future enhance concealed objects features in surrounding-guided way.} Here, in general, we can fuse the guidance features with initial prediction $O_c$ , and optimize final prediction $O_f$ through multi-stage refinement as follows:
\begin{small}
\begin{equation}
  O_c^{k-1}=O_c^{(k)}+G_{obj}^{(k)}-G_{sur}^{(k)},
\end{equation}
\end{small}

\noindent where the guidance features are composed of object texture features $G_{obj}^{(k)}$ and surrounding-aware features $G_{sur}^{(k)}$ from different layer $k\in{4, 3, 2}$. And hence, we can compute final prediction $O_f=O_c^{(1)}$.

In addition, we can use group-reversal attention~\cite{fan2021concealed}, to make object texture features reveal the potential region of concealed objects. However, it meanwhile can make some background pixels mis-predicted. To address such issue, we introduce surrounding-aware features to mine out the remaining concealed details via erasing the surrounding area, as shown in Fig.~\ref{fig3}. Specifically, surrounding-aware features $G_{sur}$ can be learned through the following process. We first obtain surrounding constraint map $C_{sur}$ from surrounding map $O_{sur}$ by Sigmoid function. Then, fuse the texture features and edge features, which are learned from the encoder, respectively. Finally, via group guidance operation, to the fusion features beneath the surrounding constraint map can be enhanced easily. Thus, through the surrounding fusion layer by layer, SurANet can achieve a final precise results. The above process can be formulated as follows:
\begin{small}
\begin{equation}
  G_{sur}=E^{GGO}(C_{sur}, F_{fusion}),
\end{equation}
\end{small}
\begin{small}
\begin{equation}
  C_{sur}=\frac{1}{e^{-O_{sur}}+1},
\end{equation}
\end{small}
\begin{small}
\begin{equation}
  F_{fusion}=BasicConv(\oplus(F_t, F_e)).
\end{equation}
\end{small}

\noindent Here, $E^{GGO}$ denotes the group guidance operation. $F_t\in \mathbb{R}^{H\times W\times C}$ and $F_e\in \mathbb{R}^{H\times W\times C}$ are first concatenated by $\oplus$ in channel dimension, and then convolved to obtain $F_{fusion}\in \mathbb{R}^{H\times W\times C}$.

\subsection{Surrounding-Aware Contrastive Loss}
\label{subsec:SACLoss}

\begin{figure}[tb!]
  \setlength{\abovecaptionskip}{0cm} 
  \setlength{\belowcaptionskip}{-0.2cm}
  \centering
  \includegraphics[width=85mm]{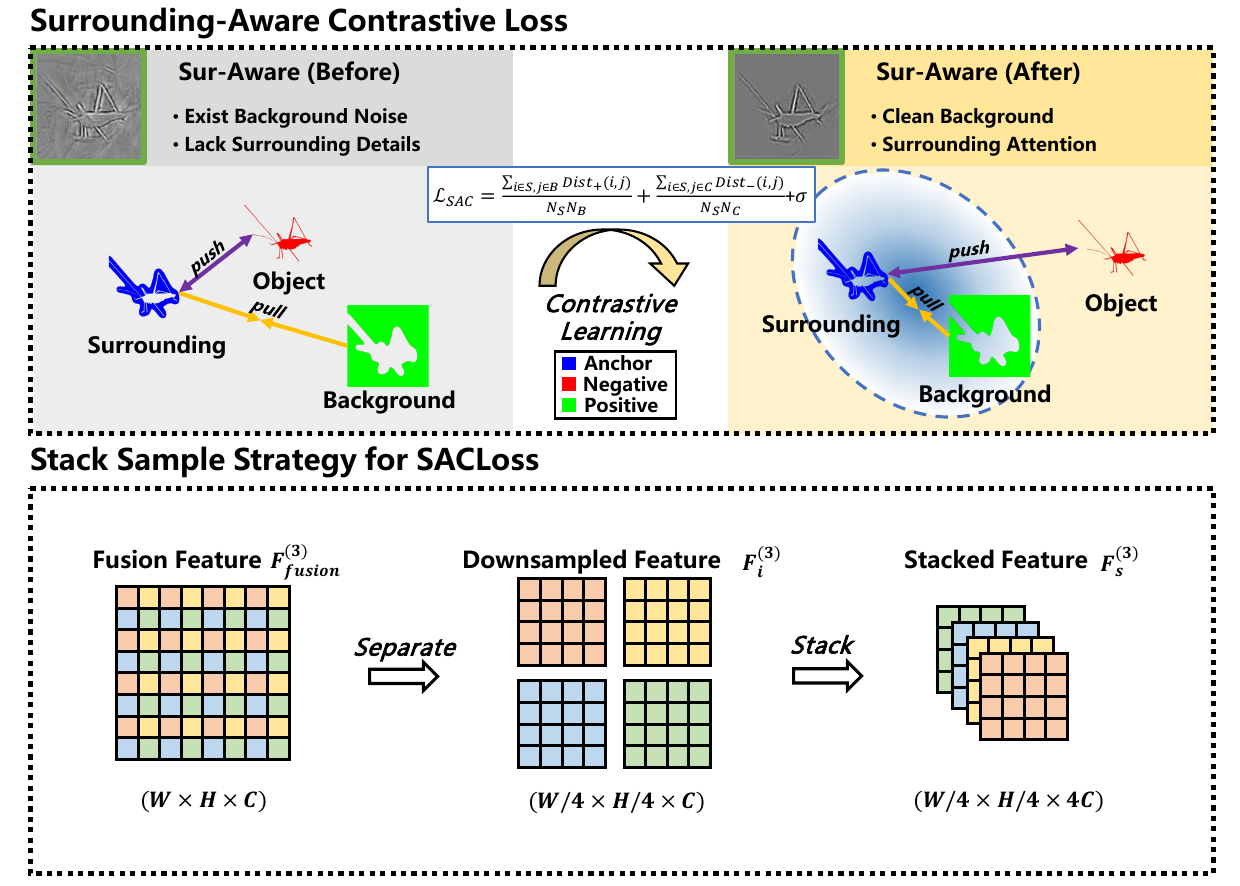}
  \caption{\textrm{\textcolor{black}{The proposed SACLoss and Spatial-Compressed Correlation Transmission (SCCT) strategy. SACLoss pushes away the negative sample pair formed by the pixels of the object/surrounding area, and pulls back the positive sample pair formed by the pixels of the background/surrounding area, respectively. After training process, the difference between object and background is amplified, which makes surrounding-aware feature (Sur-Aware) more obvious. The SCCT strategy transform feature maps into compact representations and enables SACLoss to efficient capture the relationships. It's divided into separation and stack operation. In this case, the number of layers $k$ is 3. Fusion features $F^{(3)}_{fusion}$ are divided into four independent parts $F^{(3)}_i$ by interval, and stack them according to the channel dimension.}}}
  \label{fig5}
\end{figure}

In this subsection, we will introduce the detail of Surrounding-Aware Contrastive Loss (SACLoss). 

Recent experiments have already demonstrated the effectiveness of exploiting local or global contextual information in individual image and supervising the model with cross-entropy loss~\cite{pang2022zoom, li2021uncertainty}. 
However, since concealed objects can create spurious semantic relations, it becomes more difficult to discriminate objects from the surrounding area. 

Contrastive loss is a better way to enhance the similarity of semantically similar pixels while maintaining the distinction from other pixels~\cite{wang2021understanding}, giving competitive classification performance. Consequently, in order to further differential surrounding features and concealed object features, we proposed the Surrounding-Aware Contrastive Loss (SACLoss). Here, SACLoss can enlarge the distance between surrounding features and the object's, and reduce the distance between surrounding features and the background's. Via such contrastively learning strategy, the feature map of concealed objects can be better identified.

Formally, let $f_i$ and $f_j$ $\in\mathbb{R}^{C}$ be the sample feature from fusion feature $F_{fusion}\in\mathbb{R}^{H\times W\times C}$, respectively, where $i$ and $j$ $\in \{ 1, ..., H\times W \}$, $i\neq j$. $f_i$ and $f_j$ constitute a positive sample pair $pair_+(f_i, f_j)$ if they represent same classes, where the distance metric $Dist_+(i, j)$ can be expressed as follows:
\begin{small}
\begin{equation}
  Dist_+(i, j)=-\left\lVert f_i, f_j \right\rVert _2.
\end{equation}
\end{small}

\noindent Otherwise, they constitute a negative sample pair $pair_-(f_i, f_j)$, and distance metric $Dist_-(i, j)$ can be expressed as:
\begin{small}
\begin{equation}
  Dist_-(i, j)=\left\lVert f_i, f_j \right\rVert _2.
\end{equation}
\end{small}

Therefore, a set of sample pairs can be constructed from $ F_{fusion}$. We regard the sample features $f_s \in S$ of the surrounding area as anchors, and form a positive sample pair $pair_+(f_s, f_b)$ with background feature $f_b$ and a negative sample pair $pair_-(f_s, f_c)$ with concealed object texture feature $f_c$. And finally, Surrounding-Aware Contrastive Loss $\mathcal{L}_{SAC}$ can be computed as follows:
\begin{small}
\begin{equation}
\begin{split}
  \mathcal{L}_{SAC}=&\frac{\sum_{i \in S, j \in B}Dist_+(i, j)}{N_SN_B}\\
  &+\frac{\sum_{i \in S, j \in C}Dist_-(i, j)}{N_SN_C}+\sigma,
\end{split}
\end{equation}
\end{small}

\noindent where $S$, $B$, $C$ denote different areas from the surrounding background and object, respectively. $N_S$, $N_B$, $N_F$ denote the number of sample features from the surrounding background and object, respectively.

\subsection{Spatial-Compressed Correlation Transmission Optimization}
\textcolor{black}{During the training process, Ground Truth, Edge Label, and Surrounding Label are required as our supervisory information. The network updates its parameters by calculating coarse loss $\mathcal{L}_{coarse}$, prediction loss $\mathcal{L}_{pred}$, and Surrounding contrast loss $\mathcal{L}_{SAC}$. The total joint-loss function can be expressed as:}
\begin{small}
	\begin{equation}
		\mathcal{L} = \mathcal{L}_{coarse} + \mathcal{L}_{pred}+ \mathcal{L}_{SAC}.
	\end{equation}
\end{small}

\noindent Here, the loss function is composed of three parts, and $\mathcal{L}_{coarse}$ contains the cross-entropy loss of three tasks in the first stage. $\mathcal{L}_{pred}$ contains the cross-entropy loss of each layer's prediction result. For not introducing much extra significant computational costs, we propose SCCT strategy for training process. 

\textcolor{black}{Generally, feature maps in the shallower layers of the network contain more object details. However, as the size of feature maps becomes larger, it gradually becomes difficult to compute from all samples in the $\mathcal{L}_{SAC}$. Due to the huge amount of calculation, we design a reasonable and efficient sampling strategy (named Spatial-Compressed Correlation Transmission) to optimize the calculation of $\mathcal{L}_{SAC}$ as shown in Fig.~\ref{fig5}, which strategy can transform feature maps into compact representations while preserving spatial correlations among pixels. In this way, the potential sample space required for calculating $\mathcal{L}_{SAC}$ is greatly reduced, and the key information of surrounding features are fully preserved. This enables $\mathcal{L}_{SAC}$ to quickly capture the relationships of different objects on extensive feature levels.} 

Specifically, for features $F_{fusion}^{(k)}$ in layer $k$, we divide it equally according to the following rules:
\begin{small}
	\begin{equation}
		F_i^{(k)}=F_{fusion}^{(k)}[C, H_i, W_i],
	\end{equation}
\end{small}

\noindent where $H_i$ and $W_i$ $\in \{i, i+(-k+5), ..., i+n \times (-k+5)\}$, and $i\in \{1, ..., (-k+5)\}$. Then, we stack all $F_i^{(k)}$ in channel dimension as follows:
\begin{small}
	\begin{equation}
		F_s^{(k)}=\oplus(F_1^{(k)}, ..., F_{-k+5}^{(k)}),
	\end{equation}
\end{small}

\noindent where $F_s^{(k)} \in\mathbb{R}^{C(-k+5)\times \frac{H}{-k+5}\times \frac{W}{-k+5}}$. Subsequently, With this approach, we can compute $Loss_{SAC}$ on shallow feature maps.

Finally, we demonstrate the optimized training process via SCCT as shown in Algorithm~\ref{alg1}, and in inference process, SurANet does not require any labelled information to segment concealed objects through training SAE module. Thus, the network can enhance the discrimination of concealed objects and surrounding area by amplifying the feature differences, thereby achieving more segmentation precise, and the underlying model architecture can still remain consistent during the training process.
\begin{algorithm}[tb!]
	\setlength{\abovecaptionskip}{0cm} 
	\setlength{\belowcaptionskip}{-0.2cm}
	\caption{Optimized Training Process via Spatial-Compressed Correlation Transmission}
	\label{alg1}
	\begin{algorithmic}
		\REQUIRE
		Input Image $I\in\mathbb{R}^{3\times H\times W}$,
		Ground Truth $GT\in\mathbb{R}^{H\times W}$,
		Edge Label $Edge\in\mathbb{R}^{H\times W}$,
		\WHILE{$t<E_{max}$}
		\STATE 1. \textcolor{black}{Generate surrounding label $Lm$ by $GT$.}
		\STATE 2. Predict coarse concealed map $O_c$, edge map $O_{edge}$ and surrounding map $O_{sur}$ on the first stage.
		\STATE 3. Compute coarse loss $\mathcal{L}_{coarse}$ on the first stage.
		\STATE 4. Enhance the texture features of each layer through group-reversal attention.
		\STATE 5. Enhance ${O_{sur}}$ with $F_c$ and $F_e$, outputs predicted image $O_f$.
		\STATE 6. Compute prediction loss $\mathcal{L}_{pred}$ for each layer on the second stage.
		\STATE 7. Compute Surrounding-Aware Contrastive Loss $\mathcal{L}_{SAC}$ with Stack Sample strategy for each layer.
		\STATE 8. Train SurANet coaching by total loss, and update SurANet parameters $\theta $.
		\ENDWHILE
	\end{algorithmic}
\end{algorithm}

\begin{table}[tb!]
\textrm{
    \setlength{\abovecaptionskip}{0cm} 
	\setlength{\belowcaptionskip}{-0.2cm}
	\centering
	\caption{\textrm{Hyper-Parameter Settings of Backbones}}
	\label{tab1}
	\renewcommand{\arraystretch}{1.5}
	\scalebox{0.75}{
		\begin{tabular}{c|c|c|c|c}
			\hline
			\textbf{Method} & \textbf{Backbone} & \textbf{Layer Size} & \textbf{Layer Channels} & \textbf{Params (M)} \\ 
			\hline
			SurANet-R & ResNet-50~\cite{he2016deep}     & \makecell{$[352, 88,$\\$44, 22, 11]$}	&\makecell{$[3,128,256,$\\$1024, 2048]$}	& 25.56 \\
			SurANet-C & ConvNext-Tiny~\cite{liu2022convnet} &	\makecell{$[352, 88,$\\$44, 22, 11]$}	&\makecell{$[3,96,192,$\\$384, 768]$}& 28.59 \\
			\hline
		\end{tabular}}
}
    \end{table}

\begin{table*}[tb!]
\textrm{
    \setlength{\abovecaptionskip}{0cm} 
	\setlength{\belowcaptionskip}{-0.2cm}
	\centering
	\caption{\textrm{Comparison of SurANet with State-Of-The-Art Methods in terms of S-Measure $S_{\alpha}$~\cite{fan2017structure}, weighted F-Measure $F_{w\beta}$~\cite{margolin2014evaluate}, mean absolute error $MAE$~\cite{perazzi2012saliency}, and e-measure score $E_\phi$~\cite{fan2021cognitive}.}}
	\label{tab2}
	\renewcommand{\arraystretch}{1.3}
	\scalebox{0.93}{   
		\begin{tabular}{cl|cccc|cccc|cccc}
			\hline
			\multicolumn{2}{c}{\multirow{3}*{Method}} \vline& \multicolumn{4}{c}{COD10K} \vline& \multicolumn{4}{c}{CAMO} \vline& \multicolumn{4}{c}{CHAMELEON} \\
			{} & {} & $S_{\alpha}$ & $F_{w\beta}$ & MAE &  $E_\phi$ & $S_{\alpha}$ & $F_{w\beta}$ & MAE &  $E_\phi$ & $S_{\alpha}$ & $F_{w\beta}$ & MAE & $E_\phi$ \\
			{} & {} & $\uparrow$ & $\uparrow$& $\downarrow$ & $\uparrow$ & $\uparrow$ & $\uparrow$& $\downarrow$ & $\uparrow$ & $\uparrow$ & $\uparrow$& $\downarrow$ & $\uparrow$\\
			\hline
			\multirow{5}*{\rotatebox{270}{GOS Method}} &\cellcolor{gray!20}FPN (CVPR'17)~\cite{lin2017feature}       &\cellcolor{gray!20}0.697  &\cellcolor{gray!20}0.411  &\cellcolor{gray!20}0.075  &\cellcolor{gray!20}0.691  &\cellcolor{gray!20}0.684  &\cellcolor{gray!20}0.483  &\cellcolor{gray!20}0.131  &\cellcolor{gray!20}0.677  &\cellcolor{gray!20}0.794  &\cellcolor{gray!20}0.590  &\cellcolor{gray!20}0.075  &\cellcolor{gray!20}0.783 \\
			{} & MaskRCNN (ICCV'17)~\cite{he2017mask}      & 0.613  & 0.402  & 0.080  & 0.748  & 0.574  & 0.430  & 0.151  & 0.715  & 0.643  & 0.518  & 0.099  & 0.778 \\
			{} &\cellcolor{gray!20}UNet++ (CVPR'18)~\cite{zhou2018unet++}    &\cellcolor{gray!20}0.623  &\cellcolor{gray!20}0.350  &\cellcolor{gray!20}0.086  &\cellcolor{gray!20}0.672  &\cellcolor{gray!20}0.599  &\cellcolor{gray!20}0.392  &\cellcolor{gray!20}0.149  &\cellcolor{gray!20}0.653  &\cellcolor{gray!20}0.695  &\cellcolor{gray!20}0.501  &\cellcolor{gray!20}0.094  &\cellcolor{gray!20}0.762 \\
			{} & SwinTrans (CVPR'18)~\cite{liu2021swin}    & 0.773 & 0.647 & 0.034 & 0.830 & 0.767 & 0.700 & 0.079 & 0.829 & 0.817 & 0.740 & 0.037 & 0.884\\
			{} &\cellcolor{gray!20}ConvNext (CVPR'18)~\cite{liu2022convnet}    &\cellcolor{gray!20}0.778 &\cellcolor{gray!20}0.655 &\cellcolor{gray!20}0.034 &\cellcolor{gray!20}0.844 &\cellcolor{gray!20}0.759 &\cellcolor{gray!20}0.685 &\cellcolor{gray!20}0.085 &\cellcolor{gray!20}0.821 &\cellcolor{gray!20}0.849 &\cellcolor{gray!20}0.783 &\cellcolor{gray!20}0.033 &\cellcolor{gray!20}0.917\\
			\hline
			\multirow{14}*{\rotatebox{270}{COD Method}} & SINet (CVPR'20)~\cite{fan2020camouflaged} & 0.771  & 0.551  & 0.051  & 0.806  & 0.751  & 0.606  & 0.100  & 0.771  & 0.869  & 0.740  & 0.044  & 0.891 \\
			{} &\cellcolor{gray!20}TINet (AAAI'21)~\cite{zhu2021inferring}   &\cellcolor{gray!20}0.793  &\cellcolor{gray!20}0.635  &\cellcolor{gray!20}0.043  &\cellcolor{gray!20}0.848  &\cellcolor{gray!20}0.781  &\cellcolor{gray!20}0.678  &\cellcolor{gray!20}0.089  &\cellcolor{gray!20}0.847  &\cellcolor{gray!20}0.874  &\cellcolor{gray!20}0.783  &\cellcolor{gray!20}0.038  &\cellcolor{gray!20}0.916 \\
			{} & LSR (CVPR'21)~\cite{lv2021simultaneously} & 0.793  & 0.663  & 0.041  & 0.868  & 0.793  & 0.696  & 0.085  & 0.826  & 0.892  & 0.812  & 0.033  & 0.928 \\
			{} &\cellcolor{gray!20}PFNet (CVPR'21)~\cite{mei2021camouflaged} &\cellcolor{gray!20}0.800  &\cellcolor{gray!20}0.660  &\cellcolor{gray!20}0.040  &\cellcolor{gray!20}0.868  &\cellcolor{gray!20}0.782  &\cellcolor{gray!20}0.695  &\cellcolor{gray!20}0.085  &\cellcolor{gray!20}0.852  &\cellcolor{gray!20}0.882  &\cellcolor{gray!20}0.810  &\cellcolor{gray!20}0.033  &\cellcolor{gray!20}0.942 \\
			{} & JCOD (CVPR'21)~\cite{li2021uncertainty}   & 0.809  & 0.684  & 0.035  & 0.884  & 0.800  & 0.728  & 0.073  & 0.859  & 0.891  & 0.817  & 0.030  & 0.943 \\
			{} &\cellcolor{gray!20}C2FNet (IJCAI'21)~\cite{sun2021context}    &\cellcolor{gray!20}0.813  &\cellcolor{gray!20}0.686  &\cellcolor{gray!20}0.036  &\cellcolor{gray!20}0.890  &\cellcolor{gray!20}0.796  &\cellcolor{gray!20}0.719  &\cellcolor{gray!20}0.080  &\cellcolor{gray!20}0.854  &\cellcolor{gray!20}0.888  &\cellcolor{gray!20}0.828  &\cellcolor{gray!20}0.032  &\cellcolor{gray!20}0.935 \\
			{} & R-MGL (CVPR'21)~\cite{zhai2021mutual}     & 0.814  & 0.666  & 0.035  & 0.865  & 0.775  & 0.673  & 0.088  & 0.847  & 0.893  & 0.813  & 0.030  & 0.923 \\
			{} &\cellcolor{gray!20}UGTR (ICCV'21)~\cite{yang2021uncertainty} &\cellcolor{gray!20}0.818  &\cellcolor{gray!20}0.677  &\cellcolor{gray!20}0.035  &\cellcolor{gray!20}0.850  &\cellcolor{gray!20}0.785  &\cellcolor{gray!20}0.686  &\cellcolor{gray!20}0.086  &\cellcolor{gray!20}0.859  &\cellcolor{gray!20}0.888  &\cellcolor{gray!20}0.796  &\cellcolor{gray!20}0.031  &\cellcolor{gray!20}0.918 \\
			{} & SINetV2 (PAMI'22)~\cite{fan2021concealed} & 0.814  & 0.676  & 0.037  & 0.886  & 0.819  & 0.741  & 0.072  & 0.874  & 0.890  & 0.814  & 0.032  & 0.939 \\
			{} &\cellcolor{gray!20}FAPNet (TIP'22)~\cite{zhou2022feature} &\cellcolor{gray!20}0.821  &\cellcolor{gray!20}0.693  &\cellcolor{gray!20}0.038  &\cellcolor{gray!20}0.885  &\cellcolor{gray!20}0.813  &\cellcolor{gray!20}0.713  &\cellcolor{gray!20}0.078  &\cellcolor{gray!20}0.864  &\cellcolor{gray!20}0.892  &\cellcolor{gray!20}0.828  &\cellcolor{gray!20}0.029  &\cellcolor{gray!20}0.938 \\
			{} & BSANet (AAAI'22)~\cite{zhu2022can} & 0.817  & 0.697  & 0.035  & 0.887  & 0.797  & 0.718  & 0.078  & 0.849  & 0.893  & 0.832  & 0.028  & 0.942 \\
			{} &\cellcolor{gray!20}PUENet (TIP'23)~\cite{zhang2023predictive} &\cellcolor{gray!20}0.812  &\cellcolor{gray!20}0.678  &\cellcolor{gray!20}0.035  &\cellcolor{gray!20}0.886  &\cellcolor{gray!20}0.792  &\cellcolor{gray!20}0.761  &\cellcolor{gray!20}0.080  &\cellcolor{gray!20}0.857  &\cellcolor{gray!20}0.888  &\cellcolor{gray!20}0.844  &\cellcolor{gray!20}0.030  &\cellcolor{gray!20}0.943 \\
			{} & diffCOD (ECAI'23)~\cite{chen2023diffusion} & 0.812  & 0.684  & 0.036  & 0.892  & 0.795  & 0.704  & 0.082  & 0.852  & 0.893  & 0.826  & 0.030  & 0.933 \\
			{} &\cellcolor{gray!20}\textbf{SurANet-R (Ours)} &\cellcolor{gray!20}\textbf{0.824}&\cellcolor{gray!20}\textbf{0.703}&\cellcolor{gray!20}\textbf{0.033}&\cellcolor{gray!20}\textbf{0.893} &\cellcolor{gray!20}\textbf{0.835}&\cellcolor{gray!20}\textbf{0.747}&\cellcolor{gray!20}\textbf{0.063}&\cellcolor{gray!20}\textbf{0.880} &\cellcolor{gray!20}\textbf{0.895}&\cellcolor{gray!20}\textbf{0.839}&\cellcolor{gray!20}\textbf{0.027}&\cellcolor{gray!20}\textbf{0.946}\\
			{} & \textbf{SurANet-C (Ours)} & \textbf{0.839}& \textbf{0.731}& \textbf{0.030}& \textbf{0.911} & \textbf{0.838}& \textbf{0.779}& \textbf{0.060}& \textbf{0.900} & \textbf{0.897}& \textbf{0.841}& \textbf{0.027}& \textbf{0.953}\\
			\hline
		\end{tabular}
	}}
\end{table*}

\begin{table*}[tb!]
\textrm{
    \setlength{\abovecaptionskip}{0cm} 
	\setlength{\belowcaptionskip}{-0.2cm}
	\centering
	\caption{\textrm{The Ablation of Proposed Modules on Three COD Datasets.}}
	\label{tab3}
	\renewcommand{\arraystretch}{1.3}
	\scalebox{0.8}{
		\begin{tabular}{c|c|c|c|cccc|cccc|cccc}
			\hline
			\multirow{3}*{No.} & \multirow{3}*{Backbone} & \multirow{3}*{Modules Setting} & \multirow{3}*{Loss Setting} & \multicolumn{4}{c}{COD10K} \vline& \multicolumn{4}{c}{CAMO} \vline& \multicolumn{4}{c}{CHAMELEON} \\
			{} & {} & {} & {} & $S_{\alpha}$ & $F_{w\beta}$ & MAE &  $E_\phi$ & $S_{\alpha}$ & $F_{w\beta}$ & MAE &  $E_\phi$ &  $S_{\alpha}$ & $F_{w\beta}$ & MAE & $E_\phi$\\
			{} & {} & {} & {} & $\uparrow$ & $\uparrow$ & $\downarrow$ & $\uparrow$ & $\uparrow$ & $\uparrow$ & $\downarrow$ & $\uparrow$ & $\uparrow$ & $\uparrow$ & $\downarrow$ & $\uparrow$ \\
			\hline
			1 & \multirow{3}*{ResNet-50} &\cellcolor{gray!20}Baseline &\cellcolor{gray!20}BCE &\cellcolor{gray!20}0.814 &\cellcolor{gray!20}0.676 &\cellcolor{gray!20}0.037 &\cellcolor{gray!20}0.886 &\cellcolor{gray!20}0.819 &\cellcolor{gray!20}0.741 &\cellcolor{gray!20}0.072 &\cellcolor{gray!20}0.874 &\cellcolor{gray!20}0.890 &\cellcolor{gray!20}0.814 &\cellcolor{gray!20}0.032 &\cellcolor{gray!20}0.939 \\
			2 & {} & +SAE & BCE & 0.817 & 0.691 & 0.034 & 0.889 & 0.831 & 0.742 & 0.065 & 0.878  & 0.893 & 0.837 & 0.029 & 0.945\\
			3 & {} &\cellcolor{gray!20}+SAE &\cellcolor{gray!20}+SAC &\cellcolor{gray!20}\textbf{0.824} &\cellcolor{gray!20}\textbf{0.703} &\cellcolor{gray!20}\textbf{0.033} &\cellcolor{gray!20}\textbf{0.893} &\cellcolor{gray!20}\textbf{0.835} &\cellcolor{gray!20}\textbf{0.747} &\cellcolor{gray!20}\textbf{0.063} &\cellcolor{gray!20}\textbf{0.880}   &\cellcolor{gray!20}\textbf{0.895} &\cellcolor{gray!20}\textbf{0.839} &\cellcolor{gray!20}\textbf{0.027} &\cellcolor{gray!20}\textbf{0.946}  \\
			\hline
			4 & \multirow{3}*{ConvNext-Tiny} & Baseline & BCE & 0.819 & 0.695 & 0.034 & 0.893 & 0.819 & 0.747 & 0.069 & 0.878  & 0.893 & 0.831 & 0.028 & 0.943 \\
			5 & {} &\cellcolor{gray!20}+SAE &\cellcolor{gray!20}BCE &\cellcolor{gray!20}0.832 &\cellcolor{gray!20}0.717 &\cellcolor{gray!20}0.031 &\cellcolor{gray!20}0.902 &\cellcolor{gray!20}0.824 &\cellcolor{gray!20}0.757 &\cellcolor{gray!20}0.064 &\cellcolor{gray!20}0.882  &\cellcolor{gray!20}0.898 &\cellcolor{gray!20}0.840 &\cellcolor{gray!20}0.027 &\cellcolor{gray!20}\textbf{0.955}\\
			6 & {} & +SAE & +SAC & \textbf{0.839} & \textbf{0.731} & \textbf{0.030} & \textbf{0.911} & \textbf{0.838} & \textbf{0.779} & \textbf{0.060} & \textbf{0.900}   & \textbf{0.897} & \textbf{0.841} & \textbf{0.027} & 0.953  \\
			\hline
		\end{tabular}
	}}
\end{table*}

\begin{figure*}[b!]
	\setlength{\abovecaptionskip}{0cm} 
	\setlength{\belowcaptionskip}{-0.2cm}
	\centering
	\includegraphics[width=170mm,height=40mm]{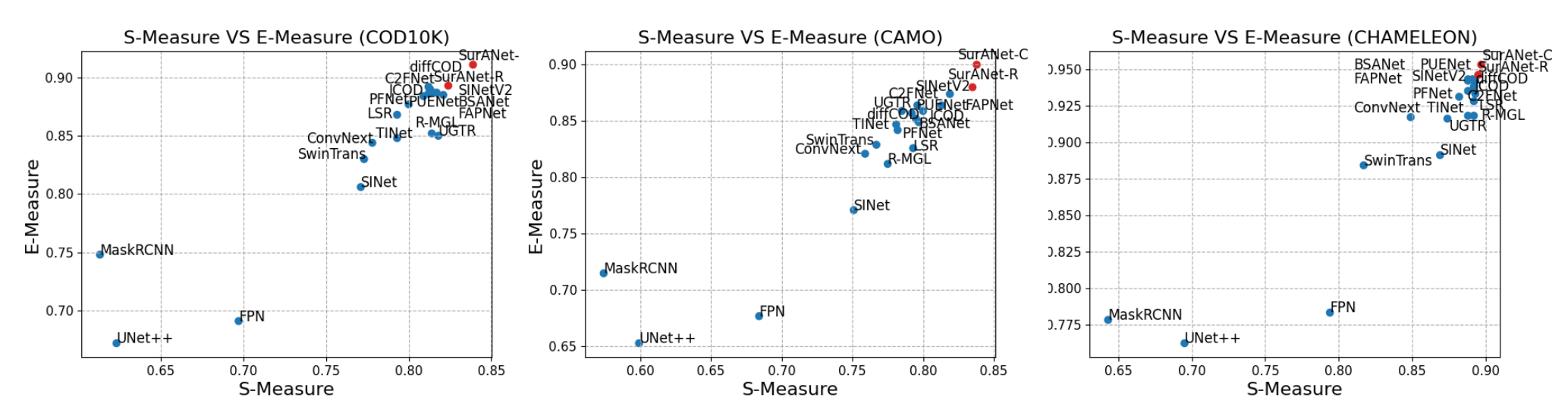}
	\caption{\textrm{S-measure $S_\alpha$ and E-measure $E_\phi$ comparison with the state-of-the-art methods. Our SurANet significantly outperforms others.}}
	\label{fig6}
\end{figure*}

\section{Experiments}
\label{sec:experiments}
In this section, we conduct extensive experiments on typical COD datasets and in several real scenes. 
First, we provide the experimental setup in section IV-A. 
Next, we compare SurANet with other state-of-the-art methods on three COD datasets in section IV-B. 
Then, we conduct experiments to evaluate the effectiveness of each component of SurANet in section IV-C. 
Finally, we evaluated SurANet via the generalization experiment of real scenes in section IV-D.

\subsection{Experimental Setup}
\textbf{Datasets}. CHAMELEON dataset was introduced by Skurowski et al.~\cite{skurowski2018animal} with only 76 images. It mostly contains some naturally concealed animals, which are collected on the Internet.
CAMO dataset was introduced by Le et al.~\cite{le2019anabranch}. It consists of 1,000 training images and 250 testing images, which contains two categories, naturally concealed animals and artificially concealed persons. 
COD10K dataset~\cite{fan2020camouflaged} is a large-scale COD benchmark with 10,000 images and covering over 78 categories of camouflaged objects in various natural scenes. 
All above datasets have corresponding pixel-level labels, considered as benchmarks for most COD tasks, and our experiments are based on these datasets.

\textbf{Implementation Details}. The implementation of our proposed method is based on PyTorch framework, using two NVIDIA GeForce RTX 2080 Ti GPUs for training. 
\textcolor{black}{During the training process, we select two different types of backbones for SurANet, and the hyper-parameter settings of backbones are listed in Table~\ref{tab1}. Here, during surrounding generation progress, we set the standard deviation of Gaussian kernel $\sigma$ as 50.} 
Furthermore, all experiments are conducted using Adam Optimizer with parameters $\beta_1 = 0.9$, $\beta_2 = 0.999$. 
The learning rate for the $t$-th epoch is $lr_0 \times d^{\frac{t}{E_d}}$, where initial learning rate $lr_0=10^{-4}$ and decay rate $d=0.1$.
Learning decay $E_d$ lasts for 30 epochs, and training lasts for a total of 100 epochs. 
All input images are resized into $352 \times 352$, and set batch is set size to 18.

\textbf{Evaluation Criteria.}
We evaluate SurANet against state-of-the-art methods and adopt 4 widely-used metrics to analyze the proposed method, i.e., mean absolute error $MAE$~\cite{perazzi2012saliency}, weighted F-measure score $F_{w\beta}$~\cite{margolin2014evaluate}, E-measure score $E_\phi$~\cite{fan2021cognitive}, and S-measure score $S_\alpha$~\cite{fan2017structure}. 
Among them, we utilize $MAE$ to assess the presence of errors, namely as follows:
\begin{small}
\begin{equation}
MAE(O,GT)=\frac{1}{HW}\sum_{i=1}^{H}\sum_{j=1}^{W}(|O(i, j)-GT(i, j)|).
\end{equation}
\end{small}
\noindent Here, $F_{w\beta}$ is used as classification-based performance evaluation, which defines the weighted Precision to measure exactness and defines the weighted Recall to measure completeness, respectively. 
It provides more reliable evaluation results than traditional $F_\beta$~\cite{achanta2009frequency}. The formula is expressed as follows:
\begin{small}
\begin{equation}
  F_{w\beta}=(1+\beta^2)\frac{{\rm Precision}^w\cdot {\rm Recall}^w}{\beta^2 \cdot {\rm Precision}^w + {\rm Recall}^w}.
\end{equation}
\end{small}
\noindent Here, $E_\phi$ can simultaneously evaluates the pixel-level matching and image-level statistics, reflecting the overall and local accuracy of concealed object detection results. $S_\alpha$ can evaluate the structural similarity of region and concealed objects.

\begin{table*}[tb!]
	\textrm{
		\setlength{\abovecaptionskip}{0cm} 
		\setlength{\belowcaptionskip}{-0.2cm}
		\centering
		\caption{\textrm{Comparison of the Surrounding-Aware Enhancement Module with Other Alternatives on Three COD Datasets.}}
		\label{tab4}
		\renewcommand{\arraystretch}{1.3}
		\scalebox{0.96}{
			\begin{tabular}{c|c|cccc|cccc|cccc}
				\hline
				\multirow{3}*{No.} & \multirow{3}*{Method} & \multicolumn{4}{c}{COD10K} \vline& \multicolumn{4}{c}{CAMO} \vline& \multicolumn{4}{c}{CHAMELEON} \\
				{} & {} & $S_{\alpha}$ & $F_{w\beta}$ & MAE &  $E_\phi$ & $S_{\alpha}$ & $F_{w\beta}$ & MAE &  $E_\phi$ &  $S_{\alpha}$ & $F_{w\beta}$ & MAE & $E_\phi$\\
				{} & {} & $\uparrow$ & $\uparrow$ & $\downarrow$ & $\uparrow$ & $\uparrow$ & $\uparrow$ & $\downarrow$ & $\uparrow$ & $\uparrow$ & $\uparrow$ & $\downarrow$ & $\uparrow$ \\
				\hline
				1 &\cellcolor{gray!20}Baseline &\cellcolor{gray!20}0.819 &\cellcolor{gray!20}0.695 &\cellcolor{gray!20}0.034 &\cellcolor{gray!20}0.893 &\cellcolor{gray!20}0.819 &\cellcolor{gray!20}0.747 &\cellcolor{gray!20}0.069 &\cellcolor{gray!20}0.878  &\cellcolor{gray!20}0.893 &\cellcolor{gray!20}0.831 &\cellcolor{gray!20}0.028 &\cellcolor{gray!20}0.943 \\
				2 & SurGenerator (Edge) & 0.819 & 0.691 & 0.035 & 0.890 & 0.810 & 0.733 & 0.074 & 0.868 & 0.888 & 0.820 & 0.032 & 0.938 \\
				3 &\cellcolor{gray!20}SurGenerator (Uni) &\cellcolor{gray!20}0.821 &\cellcolor{gray!20}0.695 &\cellcolor{gray!20}0.034 &\cellcolor{gray!20}0.892 &\cellcolor{gray!20}0.814 &\cellcolor{gray!20}0.737 &\cellcolor{gray!20}0.073 &\cellcolor{gray!20}0.871 &\cellcolor{gray!20}0.893 &\cellcolor{gray!20}0.829 &\cellcolor{gray!20}0.031 &\cellcolor{gray!20}0.941\\
				4 & SurGenerator (Gauss) & 0.823 & 0.699 & 0.034 & 0.894 & 0.816 & 0.743 & 0.072 & 0.874 & 0.895 & 0.031 & 0.030 & 0.943\\
				5 &\cellcolor{gray!20}SAE(Texture Feat) &\cellcolor{gray!20}0.836 &\cellcolor{gray!20}0.723 &\cellcolor{gray!20}0.030 &\cellcolor{gray!20}0.909 &\cellcolor{gray!20}0.836 &\cellcolor{gray!20}0.770 &\cellcolor{gray!20}0.060 &\cellcolor{gray!20}0.894  &\cellcolor{gray!20}\textbf{0.899} &\cellcolor{gray!20}\textbf{0.842} &\cellcolor{gray!20}0.026 &\cellcolor{gray!20}0.951 \\
				6 & \textbf{SAE(Fusion Feat)} & \textbf{0.839} & \textbf{0.731} & \textbf{0.030} & \textbf{0.911}  & \textbf{0.838} & \textbf{0.779} & \textbf{0.060} & \textbf{0.900} & 0.897 & 0.841 & \textbf{0.027} & \textbf{0.953} \\
				\hline
			\end{tabular}
		}}
	\end{table*}
	
	\begin{table*}[tb!]
	\textrm{
		\setlength{\abovecaptionskip}{0cm} 
		\setlength{\belowcaptionskip}{-0.2cm}    
		\centering
		\caption{\textrm{Comparison of The Surrounding-Aware Contrastive Loss with Other Alternatives on Three COD Datasets}}
		\label{tab5}
		\renewcommand{\arraystretch}{1}
		\scalebox{0.9}{
			\begin{tabular}{c|c|c|cccc|cccc|cccc}
				\hline
				\multirow{3}*{No.} & \multirow{3}*{Loss Setting} & \multirow{3}*{Sampling Setting} & \multicolumn{4}{c}{COD10K} \vline& \multicolumn{4}{c}{CAMO} \vline& \multicolumn{4}{c}{CHAMELEON} \\
				{} & {} & {} & $S_{\alpha}$ & $F_{w\beta}$ & MAE &  $E_\phi$ & $S_{\alpha}$ & $F_{w\beta}$ & MAE &  $E_\phi$ &  $S_{\alpha}$ & $F_{w\beta}$ & MAE & $E_\phi$\\
				{} & {} & {} & $\uparrow$ & $\uparrow$ & $\downarrow$ & $\uparrow$ & $\uparrow$ & $\uparrow$ & $\downarrow$ & $\uparrow$ & $\uparrow$ & $\uparrow$ & $\downarrow$ & $\uparrow$ \\
				\hline
				1 &\cellcolor{gray!20}BCE Only &\cellcolor{gray!20}- &\cellcolor{gray!20}0.832 &\cellcolor{gray!20}0.717 &\cellcolor{gray!20}0.031 &\cellcolor{gray!20}0.902 &\cellcolor{gray!20}0.824 &\cellcolor{gray!20}0.757 &\cellcolor{gray!20}0.064 &\cellcolor{gray!20}0.882  &\cellcolor{gray!20}0.898 &\cellcolor{gray!20}0.840 & \cellcolor{gray!20}0.027 & \cellcolor{gray!20}0.955\\
				2 & BCE+SAC & HighLayer & 0.831 & 0.718 & 0.031 & 0.902 & 0.828 & 0.761 & 0.063 & 0.887 & 0.896 & 0.835 & 0.029 & 0.949 \\
				3 &\cellcolor{gray!20}BCE+SAC &\cellcolor{gray!20}SubSample &\cellcolor{gray!20}0.835 &\cellcolor{gray!20}0.724 &\cellcolor{gray!20}0.031 &\cellcolor{gray!20}0.906 &\cellcolor{gray!20}0.830 &\cellcolor{gray!20}0.762 &\cellcolor{gray!20}0.064 &\cellcolor{gray!20}0.886 &\cellcolor{gray!20}\textbf{0.899} &\cellcolor{gray!20}0.841 &\cellcolor{gray!20}\textbf{0.025} &\cellcolor{gray!20}\textbf{0.956} \\
				4 & \textbf{BCE+SAC} & \textbf{SCCT} & \textbf{0.839} & \textbf{0.731} & \textbf{0.030} & \textbf{0.911} & \textbf{0.838} & \textbf{0.779} & \textbf{0.060} & \textbf{0.900}   & 0.897 & \textbf{0.841} & 0.027 & 0.953  \\
				\hline
			\end{tabular}
		}}
	\end{table*}

\subsection{Comparison with State-of-the-art Methods}
We compare SurANet with existing 18 recent regular methods on the above three different COD datasets. Here, we select 13 COD methods, SINet~\cite{fan2020camouflaged}, TINet~\cite{zhu2021inferring}, LSR~\cite{lv2021simultaneously}, PFNet~\cite{mei2021camouflaged}, JCOD~\cite{li2021uncertainty}, C2FNet~\cite{sun2021context}, R-MGL~\cite{zhai2021mutual}, UGTR~\cite{yang2021uncertainty}, SINetV2~\cite{fan2021concealed}, FAPNet~\cite{zhou2022feature}, BSANet~\cite{zhu2022can}, PUENet~\cite{zhang2023predictive}, and diffCOD~\cite{chen2023diffusion}.
Besides, to compare the differences between COD methods and general object segmentation (GOS) methods, we select 5 typical GOS methods, FPN~\cite{lin2017feature}, MaskRCNN (FPN with ROIAlign)~\cite{he2017mask}, UNet++~\cite{zhou2018unet++}, SwinTrans~\cite{liu2021swin}, and ConvNext~\cite{liu2022convnet}. 

\textcolor{black}{\textbf{Quantitative evaluations.}
We report 4 primary metrics of these methods on COD tasks in Table~\ref{tab2}. The GOS methods baseline is FPN~\cite{lin2017feature} and COD methods baseline is the SINet~\cite{fan2020camouflaged}, accordingly. By comparing these methods, we can acquire the following three conclusions.}

\textcolor{black}{1) \textbf{COD outperforms GOS.} Despite the recent improvements in GOS methods for COD task, due to the lack of specific design of concealed objects, current GOS methods are still hard to find concealed objects in various scenes. Overall, there is a noticeable gap compared to the specifically designed COD methods.}

\begin{figure*}[tb!]
	\setlength{\abovecaptionskip}{0cm} 
	\setlength{\belowcaptionskip}{-0.2cm}
	\centering
	\includegraphics[width=170mm, height=90mm]{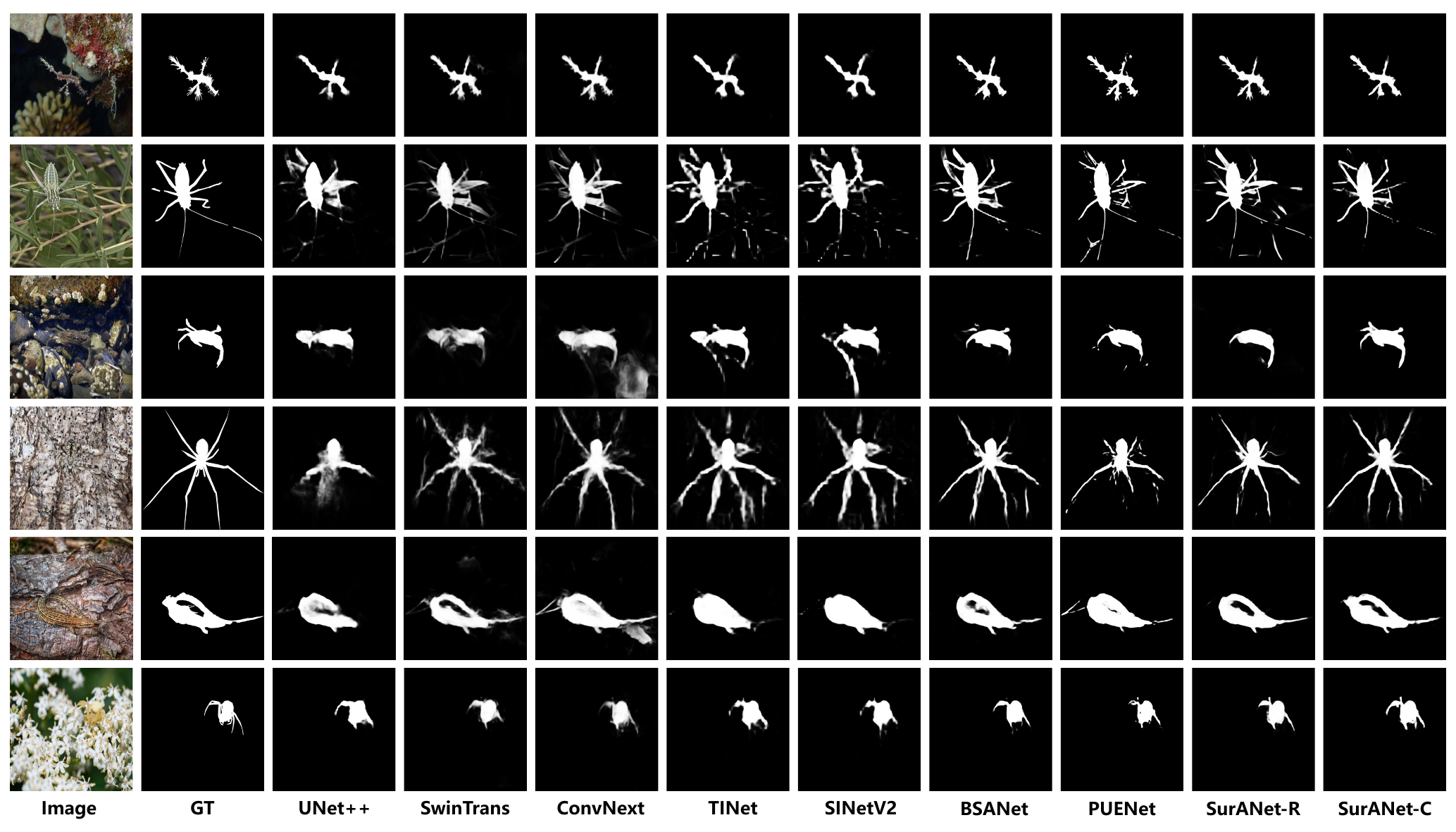}
	\caption{\textrm{Qualitative comparison of SurANet with other state-of-the-art COD methods. The first column shows 6 different kinds of concealed objects. The other columns show comparison results, and our predictions are the most similar to the ground-truth.}}
	\label{fig7}
\end{figure*}

\begin{figure}[t!]
	\setlength{\abovecaptionskip}{0cm} 
	\setlength{\belowcaptionskip}{-0.2cm}
	\centering
	\includegraphics[width=90mm]{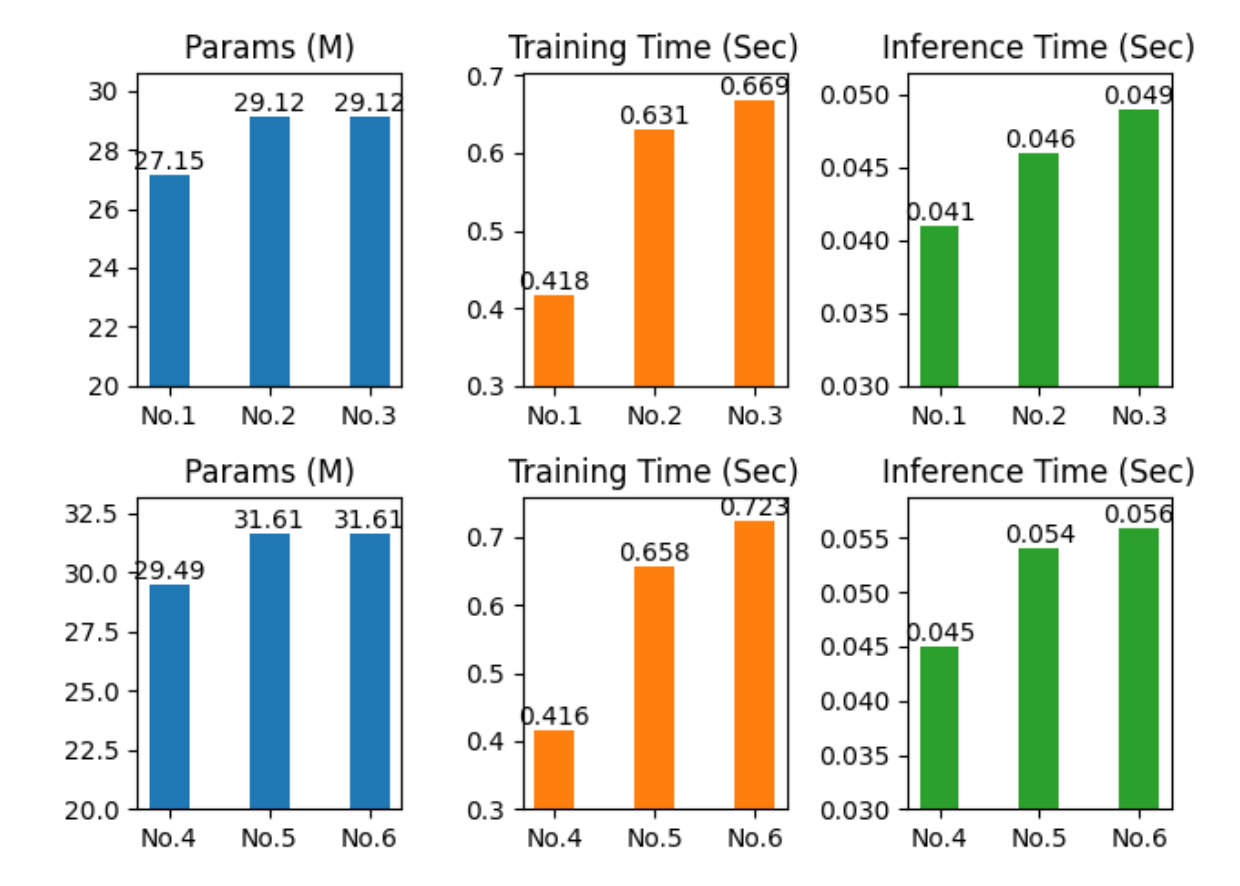}
	\caption{\textrm{Model parameters and time consumption of our proposed modules in ablation experiment. Table~\ref{tab5} reports the configuration of No.1 experiment to No.6 experiment.}}
	\label{fig8}
\end{figure}

\textcolor{black}{2) \textbf{SurANet demonstrates a better performance.} As the result of surrounding awareness, SurANet has achieved the best prediction result in object structures. Besides, experimental results on different backbones indicate that, SurANet achieves new SOTA performances across all metrics on these datasets.}

\textcolor{black}{3) \textbf{SurANet keeps both structural and accuracy.} We visualize S-measure $S_{\alpha}$ and E-measure $E_\phi$ of the result of each method in Fig.~\ref{fig6}. Through such comparison, we can conclude that, the segmentation results of SurANet can not only maintain excellent structural information, but also high global and local accuracy.}

\begin{figure*}[tb!]
	\centering
	\includegraphics[width=170mm,height=100mm]{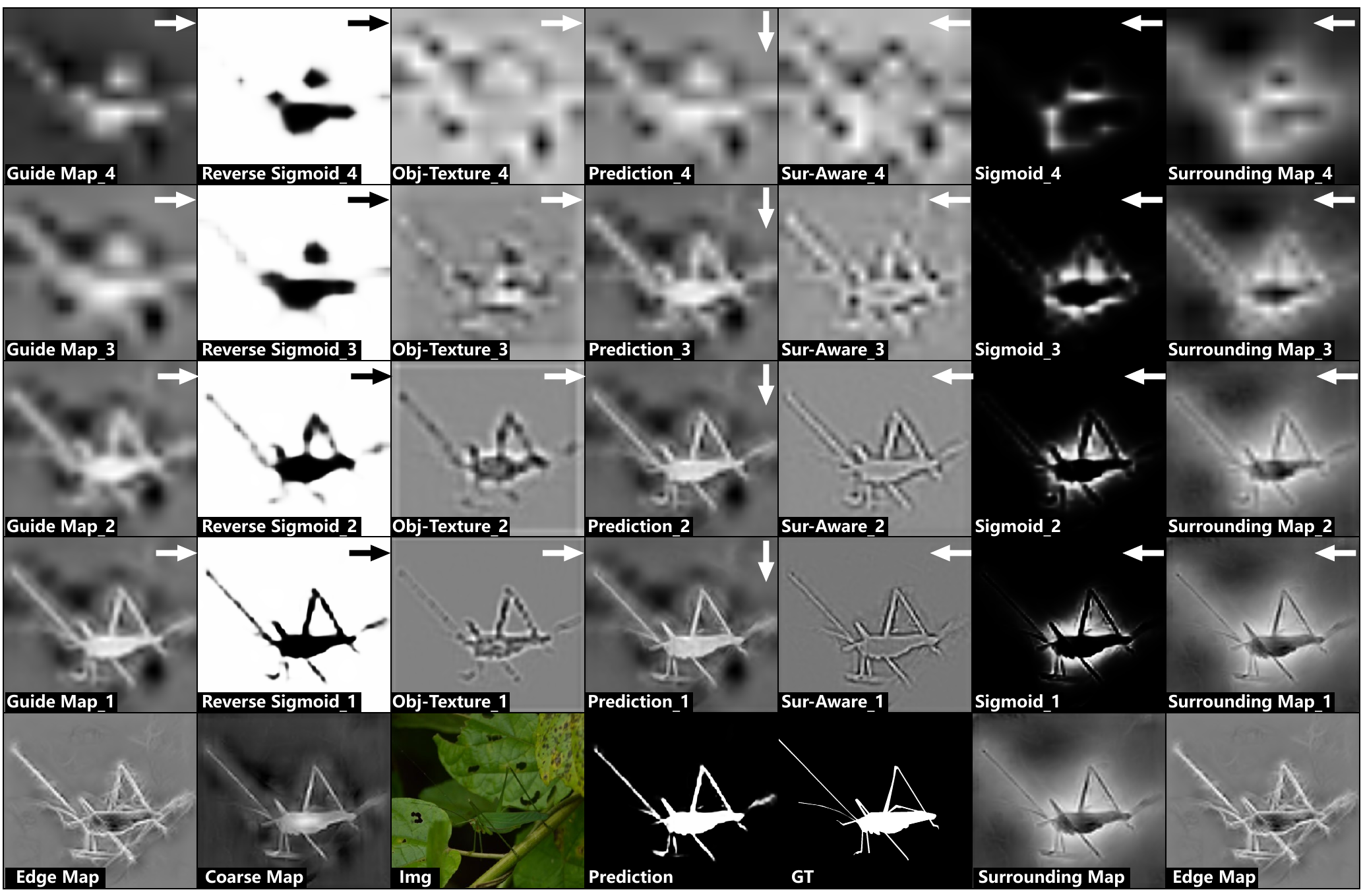}
	\caption{\textrm{Illustration of SurANet work-flow visualization. Arrows show the flow of various feature maps in the model. Except for the last row, the three columns on the left show the step by step extraction process of the concealed object feature, while the three columns on the right demonstrate the step by step extraction process of the surrounding feature. By iteratively fusing these two types of features layer by layer, the final prediction result is shown in the fourth column of the last row. The last row also shows the prediction of edge map, coarse map, surrounding map and ground truth at the same time.}}
	\label{fig9}
\end{figure*}

\begin{figure}[tb!]
	\centering
	\includegraphics[width=85mm]{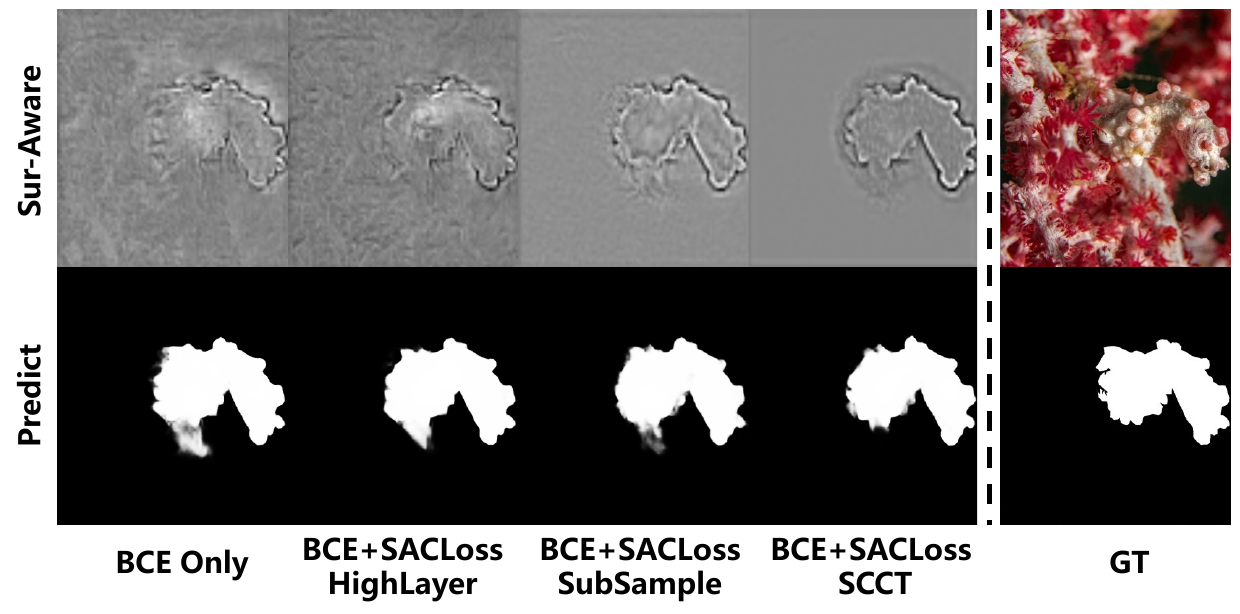}
	\caption{\textrm{Visualization of the Surrounding-Aware Contrastive Loss with other alternatives. Experimental results show that SCCT strategy can filter the environmental information more effectively.}}
	\label{fig10}
\end{figure}

\begin{figure}[tb!]
	\centering
	\includegraphics[width=85mm]{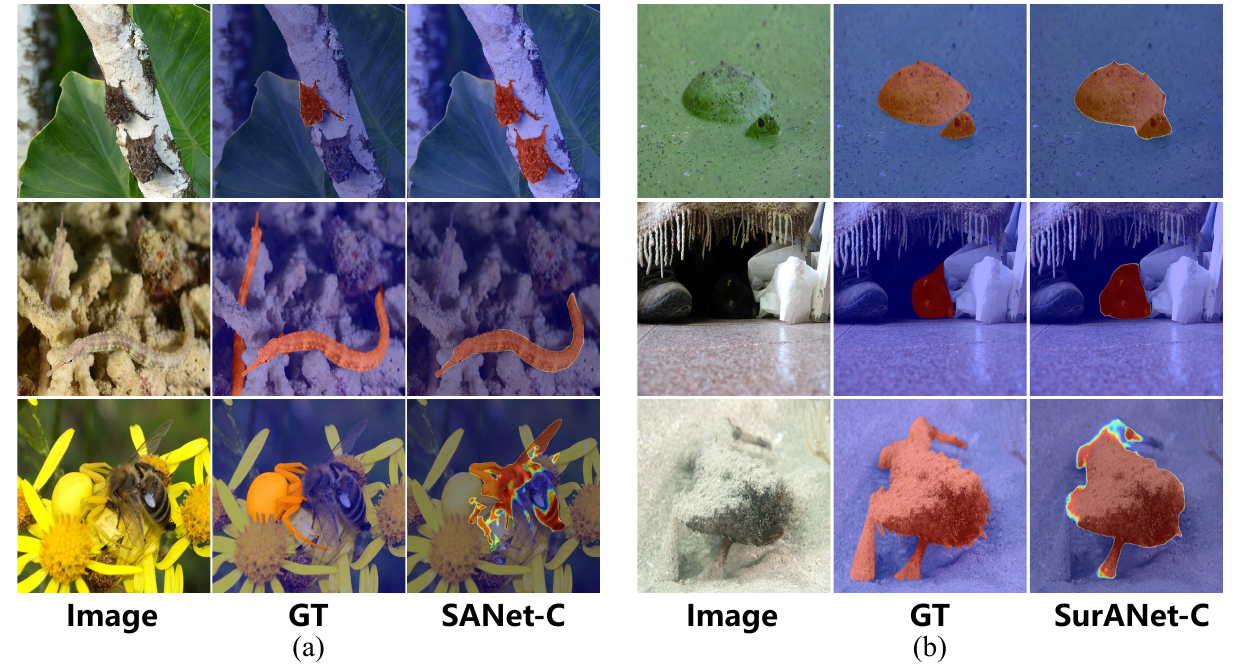}
	\caption{\textrm{Illustration of failure cases. (a) It shows some missed and false detections. (b) It shows some imperfect segmentation cases of concealed objects.}}
	\label{fig11}
\end{figure}

\textbf{Qualitative comparisons.} We provide several challenging COD results from the experiments, as shown in Fig.~\ref{fig7}, and we can further validate the conclusions drawn from quantitative evaluations, that SurANet can segment concealed objects with clear boundaries. Besides, SurANet has better discrimination in details for objects of deep concealment and complex outlines, with fewer misjudgments of background pixels.

\begin{figure*}[tb!]
	\centering
	\includegraphics[width=170mm]{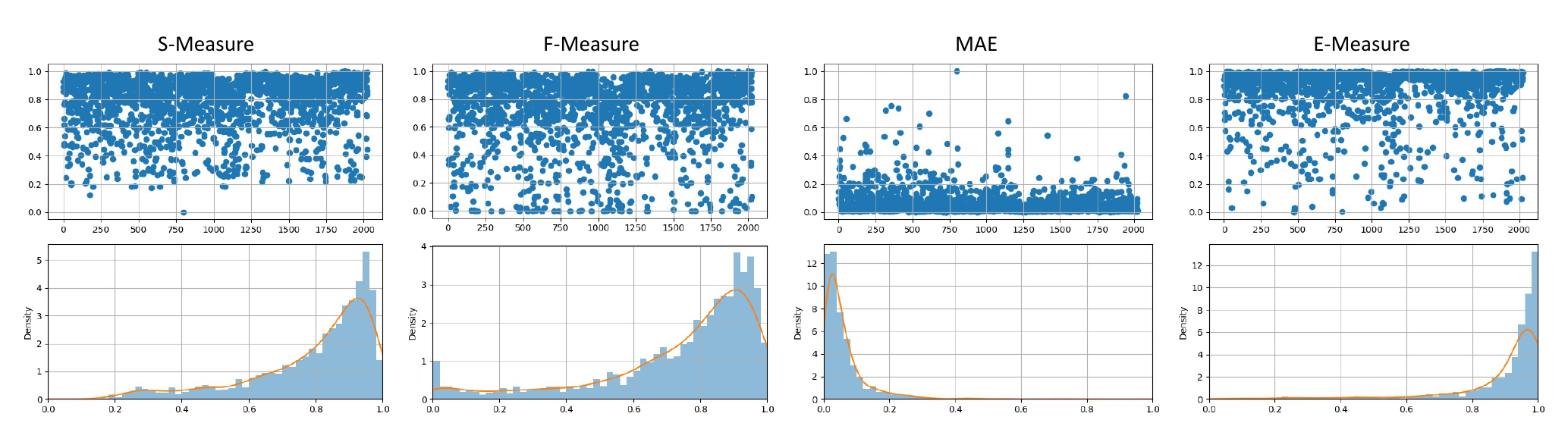}
	\caption{\textrm{Scatter plots and histograms of four evaluation metrics. The upper graphs represent the scatter plots for various metrics corresponding to each sample, and the lower ones display histograms and density estimations for these metrics.}}
	\label{fig12}
\end{figure*}

\begin{figure*}[tb!]
	\centering
	\includegraphics[width=170mm, height=35mm]{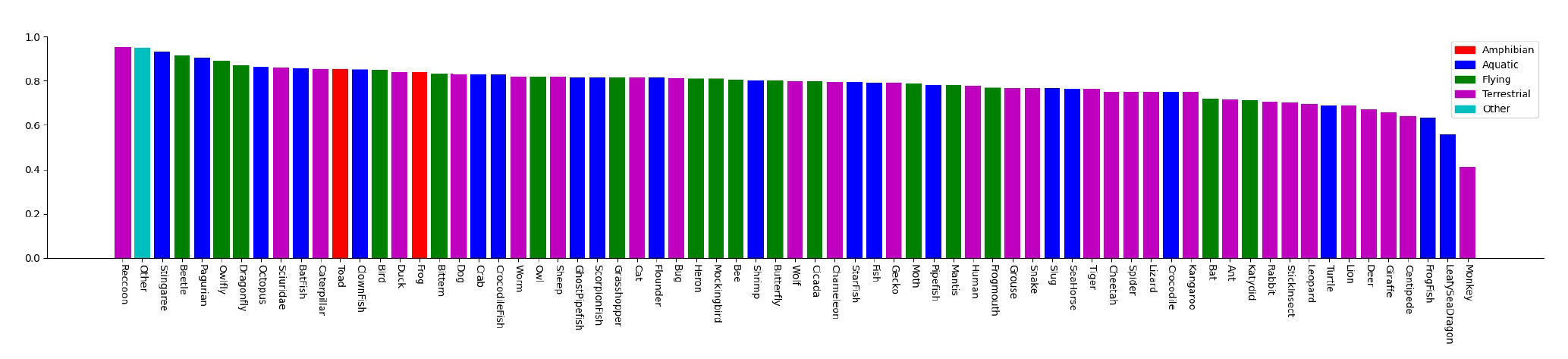}
	\caption{\textrm{S-Measure of each category concealed object in COD10K. The figure shows the S-Measure value of concealed objects from 69 subcategories within five categories, i.e., Amphibian, Aquatic, Flying, Terrestrial, and Other.}}
	\label{fig13}
\end{figure*}

\begin{figure}[tb!]
	\centering
	\includegraphics[width=85mm, height=55mm]{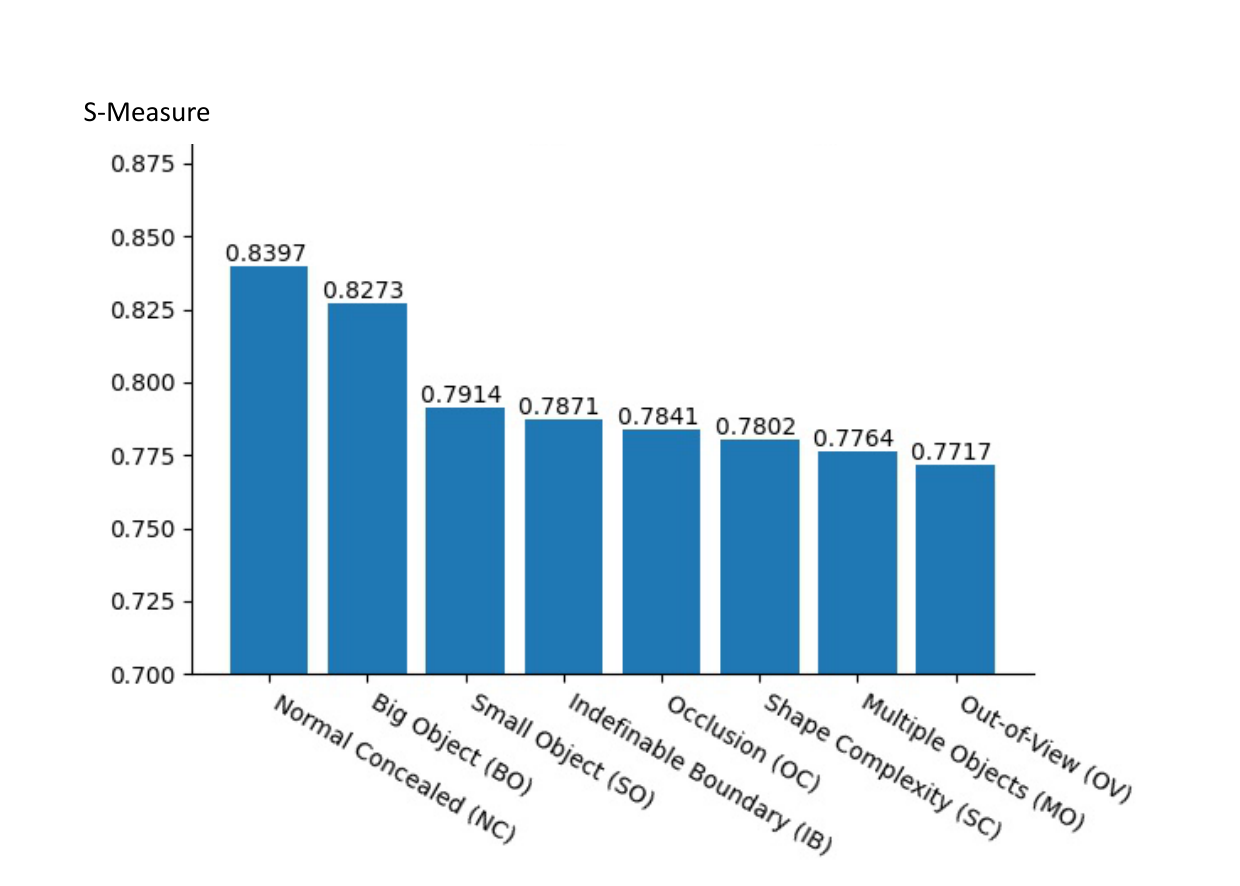}
	\caption{\textrm{S-Measure for each type of camouflage in COD10K. The x-axis displays common concealed objects from COD10K along with 7 special camouflage strategies.}}
	\label{fig14}
	\vspace{-1em}
\end{figure}

\subsection{Ablation Studies}
In this subsection, we conduct ablation study for the proposed \textcolor{black}{SAE module and SACLoss}. All experiments are based on Convnext backbone, and the settings are aforementioned in IV-A.

\textbf{Effect of our proposed modules.} 
Table~\ref{tab3} shows the results of the ablation study with SAE module and SACLoss. To facilitate a more equitable comparison, our ablation experiments were conducted using two commonly utilized backbones, ResNet-50~\cite{he2016deep} and ConvNext-tiny~\cite{liu2022convnet}. Here, we activate the 2nd to the 5th layers of SINetV2~\cite{fan2021concealed}, and use binary cross-entropy loss (BCE) training as baseline. Upon this, we make the comparison by sequentially adding SAE module and SACLoss. The experimental results demonstrate that the proposed modules maintain stable and reliable improvements across different backbones. Here, SAE module effectively enhances the performance of all metrics, and SACLoss further improves performance in most cases, which underscores the importance of surrounding awareness for COD. 

Furthermore, we evaluate the parameter amount and time consumption of our proposed module, as shown in Fig.~\ref{fig8}. Compared with the baseline, SurANet-R only increases by 1.97M, and SurANet-C only increases by 2.12M, respectively, which demonstrates uniformly positive outcomes across different backbones, with just slightly increasing the training time per image. Accordingly, thanks to SCCT strategy, SACLoss does not introduce any substantial computation costs. Hence, the inference time consumption does not increase significantly.

\textbf{Effect of various design choices of SAE.} 
Table~\ref{tab4} presents the ablation study results for the proposed SAE module. Here, SAE module synthesizes these basic components of effective and efficient surrounding feature extraction. And thus, in terms of surrounding map generation, we can compare our Generator with different definition strategies. Specifically, we select concealed object edges, uniform distribution, and Gaussian distribution to create different surrounding areas. 
The experimental results demonstrate that, the introduction of surrounding map with Gaussian distribution generation strategy achieves advantages for COD. 

Furthermore, we also compare two fusion strategies in surrounding Enhancer. 
The experimental results indicate that, integrating texture and edge feature can further enhance the surrounding awareness of SAE module. 
The visualization of each component is shown in Fig.~\ref{fig9}. 
It is clear that, as SAE module proceeds images layer by layer, the segmentation details can become more accurate.

\begin{figure*}[tb!]
	\centering
	\includegraphics[width=170mm]{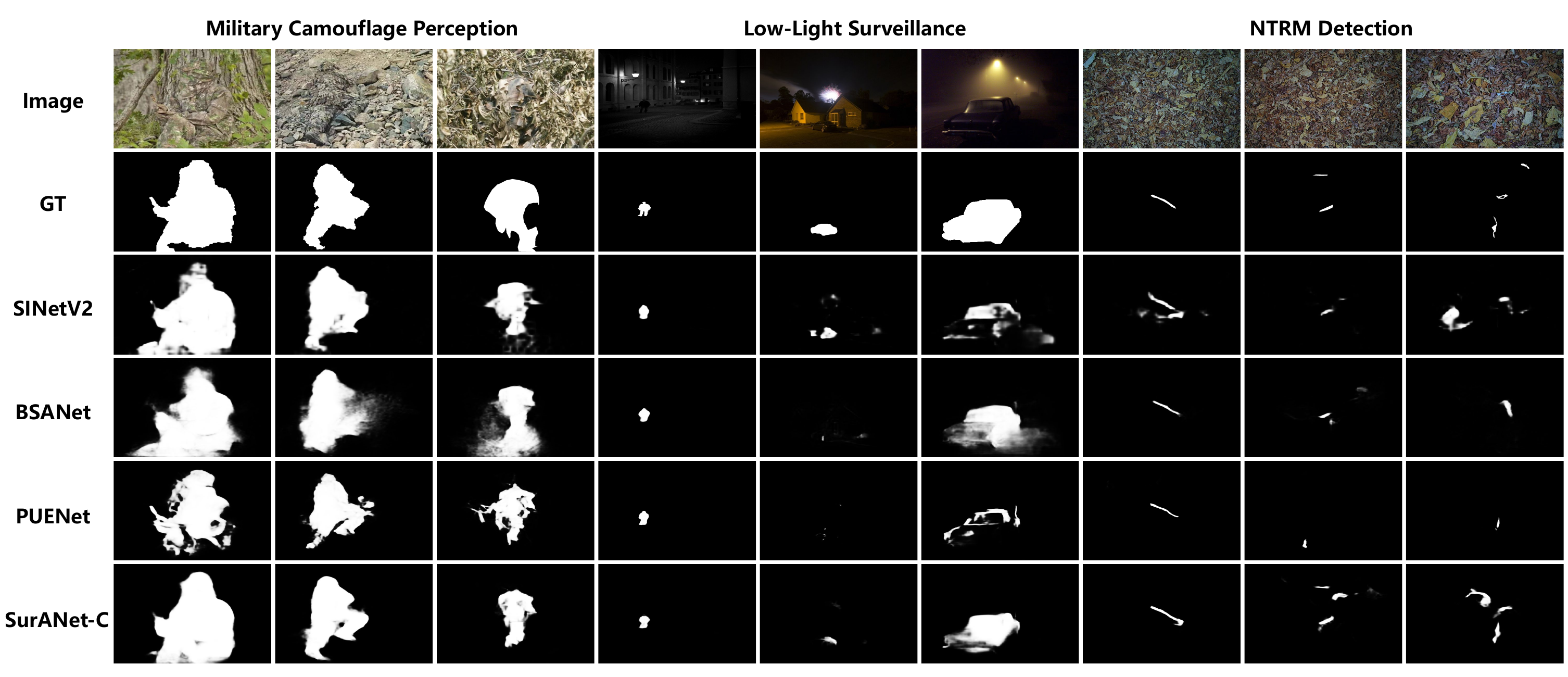}
	\caption{\textrm{Generalization application in different real-world application scenes. The first column shows different kinds of concealed objects in military camouflage perception, low-light surveillance and tobacco industry production (NTRM detection). The other columns show the comparison results. Our predictions are the most similar to the ground-truth.}}
	\label{fig15}
\end{figure*}

\textbf{Comparing SACLoss with other alternatives.} Table~\ref{tab5} shows the ablation study results of SACLoss. We try several different sampling strategies on SACLoss and compare with binary cross-entropy loss (BCE only). 
Among them, the experiment of HighLayer computes SACLoss only at the highest layer of the network, and the experiment of SubSample only computes one of the features after separation operation (as shown in Fig.~\ref{fig4}). 
Experimental results show that selecting sampling strategies can achieve the best results in most cases. 
And the visualization of each sampling strategies is shown in Fig.~\ref{fig10}. 
And thus, compared with several other strategies, SACLoss with SCCT strategy has a clearer boundary for surrounding aware features, and significantly reduces the misjudgment of the final prediction.

\textcolor{black}{\textbf{Failure Analysis.} It should be noted that, while our method significantly outperforms others in overall metrics, there are still some failure cases, as shown in Fig.~\ref{fig11}. 
Specifically, these errors can be categorized into two kinds. One is multi-object detection; the other is the detection of targets with blurred boundaries. We believe that, the main reason for these errors are the lack of sufficient training samples and overly vague boundaries of background targets, which can lead to missed and false detections, and in such challenging cases, other compared models typically yield poorer results.}

\subsection{Generalization Experiments}
Finally, we conduct generalization experiments on SurANet. Here, at first, we further evaluate the boundary performance of on COD10K. And additionally, we evaluate the generalization performance in three potential real-world application scenarios, including military camouflage perception, low-light surveillance, and tobacco industry production, which can improve the generalization ability of COD methods directly. We will next give the details of these above experiments.

\textbf{Boundary Performance.}
To analyze the boundary performance of SurANet, we conduct an analysis of the experimental results on COD10K. We first visualized the four evaluation metrics for each sample, creating scatter plots and histograms as shown in Fig.~\ref{fig12}, which reveals that most samples display a long-tail distribution in these 4 metrics, and SurANet maintains high accuracy on most samples. 
Additionally, we compare different species from 5 categories and 69 sub-categories. Here, since most samples show high consistency in the values of 4 metrics, 
we only present a comparison of S-Measure. 
As shown in Fig.~\ref{fig13}, sub-categories such as Reccon, Other, and Stingaree, have better segmentation performance, whereas FrogFish, LeafySeaDragon, and Monkey exhibit relatively poorer segmentation. 
Meanwhile, we also compare S-Measure on different special camouflage strategies of concealed objects. 
As shown in Fig.~\ref{fig14}, 
compared with normal concealed strategies, 
the other 7 special camouflage strategies have some impact on performance, especially Shape Complexity, Multiple Objects, and Out-of-view, 
as consistent with what is shown in Fig.~\ref{fig11}, which provides guidance for our further improvement of COD methods.

\textbf{Generalization Application.} We select some images for evaluation from three practical applications, including military camouflage perception, low-light surveillance, and tobacco industry production (i.e., NTRM detection task, for online non-tobacco related material detection). Here, we keep the training weights of CHAMELEON, CAMO, and COD10K dataset without any fine-tuning or re-training, and visually compare SurANet with SOTA methods directly. Experimental results are shown in Fig.~\ref{fig15}. 
For military camouflage perception, as the original training data already includes few samples, it can be noted that current methods are relatively accurate in detecting concealed military objects, and SurANet is more precise than other methods. 
For low-light surveillance, most methods can accurately segment pedestrians, but inaccurately for vehicles decreases. 
And it is obvious that, SurANet significantly outperforms other methods in vehicle segmentation. 
In NTRM detection task of tobacco industry production, images always contain mixed materials such as tobacco stems, plastic, thin ropes, feathers, etc, and we test images of extreme difficulty and find that, the NTRM segmentation accuracy of proposed SurANet is significantly superior to other methods. 
Above all, our experimental results demonstrate that our proposed SurANet has commendable generalization capabilities in various real-world COD scenes.

\section{Conclusions}
\label{sec:conclusion}
In this paper, we propose a novel Surrounding-Aware Network, namely SurANet, which introduces surrounding information into feature extraction and loss function to improve discrimination. 
In terms of extractor optimization, Surrounding-Aware Enhancement module (SAE) can enhance the semantics of feature maps by differential fusion of surrounding features into highlight concealed objects. Besides, we also introduce Surrounding-Aware Contrastive Loss (SACLoss), to better identify the concealed object, which can be efficiently trained via SCCT training strategy. 
We conduct adequate experiments on three COD datasets, CHAMELEON, CAMO, and COD10K, and three practical applications, including military camouflage perception, low-light surveillance, and tobacco industry production. Each experiment shows that the proposed SurANet outperforms state-of-the-art COD methods on multiple benchmarks~\cite{liu2020leveraging_test}. 
Moreover, considering that practical applications, many COD tasks (such as medical image disease detection) are limited by obtaining enough concealed object labels. Thus, how to efficiently learn concealed object features with less supervision information is still a major challenge~\cite{wu2022dbfnet, wu2022dca}. In the future, we plan to further study efficient concealed object feature learning methods and carry out COD tasks using weak supervised data.

\printcredits

\section*{Acknowledgments}
This work was supported in part by the National Natural Science Foundation of China under Grant 62201209, in part by Huxiang Young Talents Project Fund of Hunan Province, in part by the Open Research Fund of National Engineering Research Center for Agro-Ecological Big Data Analysis and Application, Anhui University (No. AE202207), in part by the Hunan Provincial Tobacco Company Research Project (No. HN2023KJ08), and in part by Natural Science Fund of Hunan Province under Grant 2022JJ40092.

\section*{Data Availability}
Data will be made available on request.

\bibliographystyle{mybibstyle}
\bibliography{mybibfile}

\end{document}